\documentclass[lettersize,journal]{IEEEtran}

\usepackage{amsmath,bm,bbm,amsfonts}
\usepackage{pifont}
\usepackage[table]{xcolor}
\usepackage{color}
\usepackage{soul,colortbl}

\usepackage{multirow}
\usepackage{tabularx}
\usepackage{booktabs}
\usepackage{pict2e}
\usepackage{url}
\usepackage{etoolbox}
\usepackage{orcidlink}

\usepackage{overpic}
\definecolor{cvprblue}{rgb}{0.21,0.49,0.74}
\definecolor{LightCyan}{rgb}{0.88,1,1}
\definecolor{gcolor}{RGB}{112,173,71}
\definecolor{gcolor}{RGB}{40,160,70}
\definecolor{ycolor}{RGB}{255,196,0}
\definecolor{ycolor}{RGB}{222,158,20}
\usepackage{hyperref}
\hypersetup{
     colorlinks=true,
     linkcolor=blue,
     citecolor=blue,
     filecolor=blue,
     urlcolor=magenta,
}
\usepackage[nocompress]{cite}

\usepackage{makecell}

\usepackage{todonotes}
\usepackage[nameinlink,capitalise]{cleveref}

\newtheorem{hyp}{Hypothesis}

\usepackage{subcaption}
\usepackage[switch,pagewise]{lineno}
\linenumbers

\usepackage{arydshln}
\begin{document}

\title{PHI: Bridging Domain Shift \\ in Long-Term Action Quality Assessment \\ via Progressive Hierarchical Instruction}
\author{
    Kanglei Zhou$^{\orcidlink{0000-0002-4660-581X}}$, Hubert P. H. Shum$^{\orcidlink{0000-0001-5651-6039}}$,~\IEEEmembership{Senior Member,~IEEE}, Frederick W. B. Li$^{\orcidlink{0000-0002-4283-4228}}$, \\ Xinxing Zhang$^{\orcidlink{0000-0003-4012-3796}}$, and Xiaohui Liang$^{\orcidlink{0000-0001-6351-2538}}$
    \thanks{
        Manuscript received July 16, 2024; revised March 10, 2025; accepted May 23, 2025.
        This work was supported by the National Natural Science Foundation of China under Project 62272019. \textit{(Corresponding author: Xiaohui Liang.)}
    }
    \thanks{
        Kanglei Zhou is with the State Key Laboratory of Virtual Reality Technology and Systems, Beihang University, Beijing 100191, China, and also with the Department of Computer Science, Durham University, DH1 3LE Durham, U.K. (e-mail: zhoukanglei@buaa.edu.cn).
    }
    \thanks{
        Hubert P. H. Shum and Frederick W. B. Li are with the Department of Computer Science, Durham University, DH1 3LE Durham, U.K. (e-mail: hubert.shum@durham.ac.uk; frederick.li@durham.ac.uk).
    }
    \thanks{
        Xingxing Zhang is with the Department of Computer Science and Technology, Institute for AI, BNRist Center, Tsinghua-Bosch Joint ML Center, THBI Lab, Tsinghua University. (e-mail: xxzhang1993@gmail.com).
    }
    \thanks{
        Xiaohui Liang is with the State Key Laboratory of Virtual Reality Technology and Systems, Beihang University, Beijing 100191, China, and also with the Zhongguancun Laboratory, Beijing 100190, China (e-mail: liang\_xiaohui@buaa.edu.cn).
    }
}

\markboth{IEEE Transactions on Image Processing,~Vol.~XX, No.~XX, XX~XXXX}%
{Shell \MakeLowercase{\textit{et al.}}: A Sample Article Using IEEEtran.cls for IEEE Journals}

\maketitle

\begin{abstract}
    Long-term Action Quality Assessment (AQA) aims to evaluate the quantitative performance of actions in long videos. However, existing methods face challenges due to domain shifts between the pre-trained large-scale action recognition backbones and the specific AQA task, thereby hindering their performance. This arises since fine-tuning resource-intensive backbones on small AQA datasets is impractical. We address this by identifying two levels of domain shift: task-level, regarding differences in task objectives, and feature-level, regarding differences in important features. For feature-level shifts, which are more detrimental, we propose Progressive Hierarchical Instruction (PHI) with two strategies. First, Gap Minimization Flow (GMF) leverages flow matching to progressively learn a fast flow path that reduces the domain gap between initial and desired features across shallow to deep layers. Additionally, a temporally-enhanced attention module captures long-range dependencies essential for AQA. Second, List-wise Contrastive Regularization (LCR) facilitates coarse-to-fine alignment by comprehensively comparing batch pairs to learn fine-grained cues while mitigating domain shift. Integrating these modules, PHI offers an effective solution. Experiments demonstrate that PHI achieves state-of-the-art performance on three representative long-term AQA datasets, proving its superiority in addressing the domain shift for long-term AQA.
\end{abstract}


\section{Introduction}
\IEEEPARstart{A}{ction} Quality Assessment (AQA) \cite{zhou2024magr,ji2023localization,zhang2023logo,xu2024fineparser,li2024continual} aims to evaluate the quantitative performance of actions performed in videos or image sequences.
Unlike traditional action recognition, which focuses solely on identifying specific actions, AQA provides a more detailed understanding of how well those actions are executed \cite{zhou2023hierarchical}.
This fine-grained evaluation has broad applications across domains such as sports analysis \cite{li2022pairwise,yu2021group,zhou2023hierarchical,yao2023contrastive}, medical rehabilitation \cite{zhou2023video,deb2022graph}, and skill assessment \cite{ingwersen2023video,trinh2023self,ding2023sedskill}.
Long-term AQA \cite{zeng2020hybrid,xu2022likert} extends AQA beyond individual snapshots or short clips to encompass extended durations.
This broader evaluation is more challenging yet practical, as it provides a comprehensive assessment of actions in real-world scenarios.

\begin{figure}
    \centering
    \sf
    \setlength{\tabcolsep}{0em}
    \renewcommand{\arraystretch}{1}
    \begin{tabular}{m{0.04\linewidth}<{\centering}m{0.48\linewidth}<{\centering}m{0.48\linewidth}<{\centering}}
        \cellcolor{gray!10}{\scriptsize(a)}                     & \cellcolor{orange!20}{\scriptsize Source Domain (Action Recognition)} & \cellcolor{yellow!20} \scriptsize Target Domain (AQA) \\
        \rotatebox{90}{\scriptsize \color{gray!80} Input Video} &
        \begin{overpic}[width=0.33\linewidth]{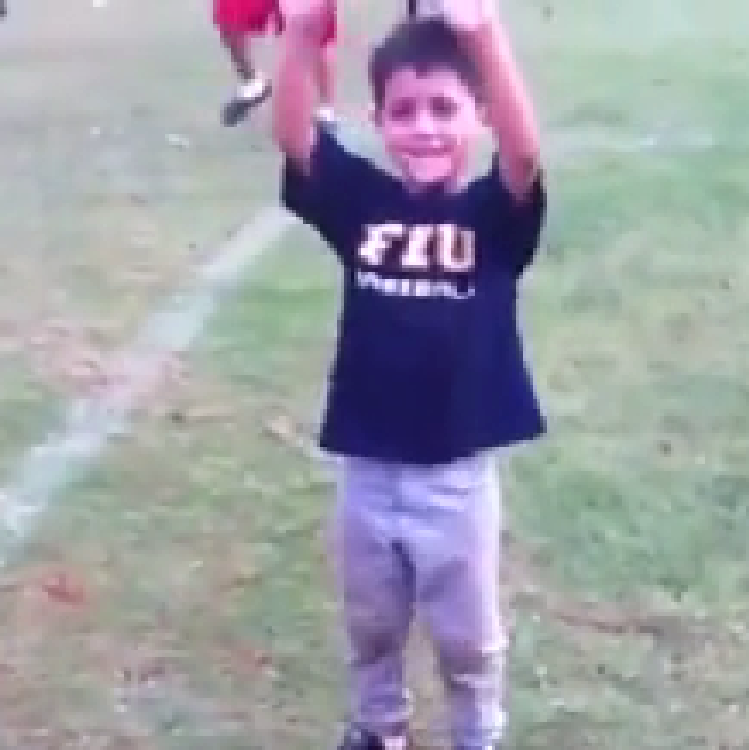}
            \put(3,5){\tiny\color{white}11${\text{-st}}$ frame}
        \end{overpic}%
        \begin{overpic}[width=0.33\linewidth]{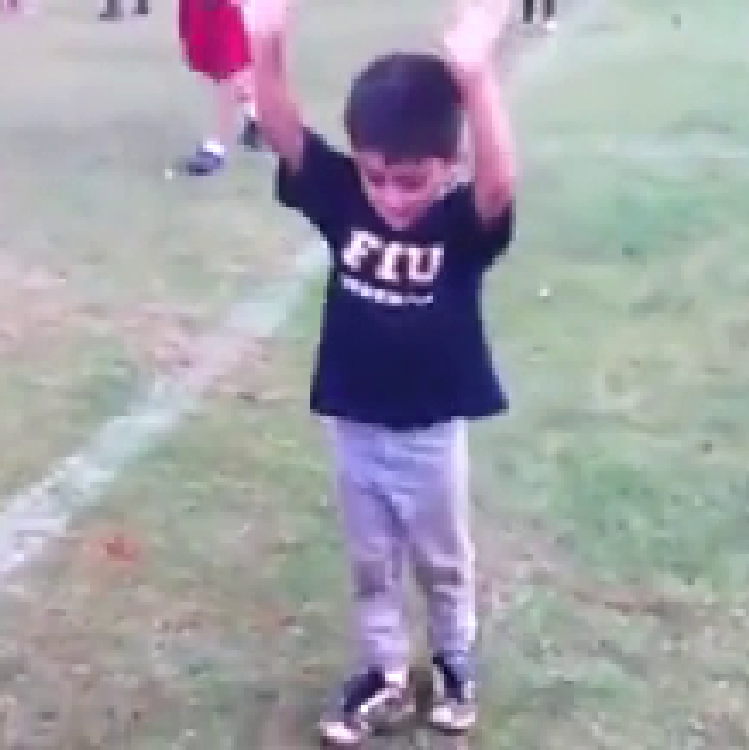}
            \put(3,5){\tiny\color{white}31${\text{-st}}$ frame}
        \end{overpic}%
        \begin{overpic}[width=0.33\linewidth]{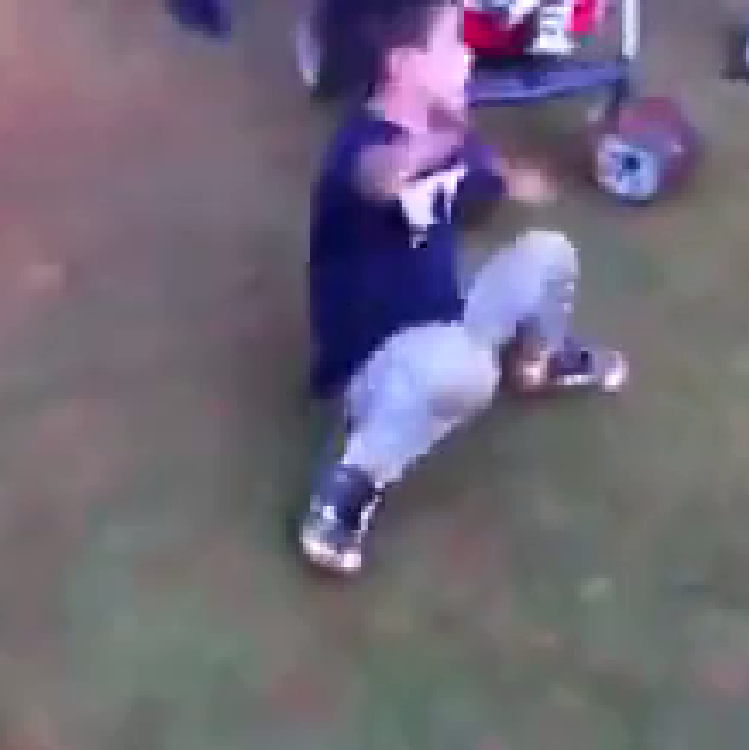}
            \put(3,5){\tiny\color{white}51${\text{-st}}$ frame}
        \end{overpic}%
                                                                &
        \begin{overpic}[width=0.33\linewidth]{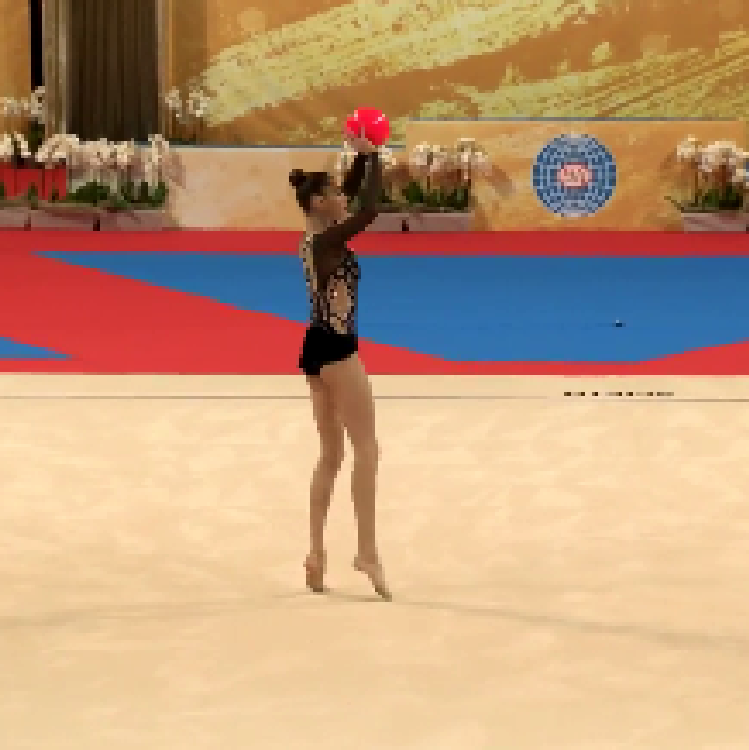}
            \put(3,5){\tiny\color{black}11${\text{-st}}$ frame}
        \end{overpic}%
        \begin{overpic}[width=0.33\linewidth]{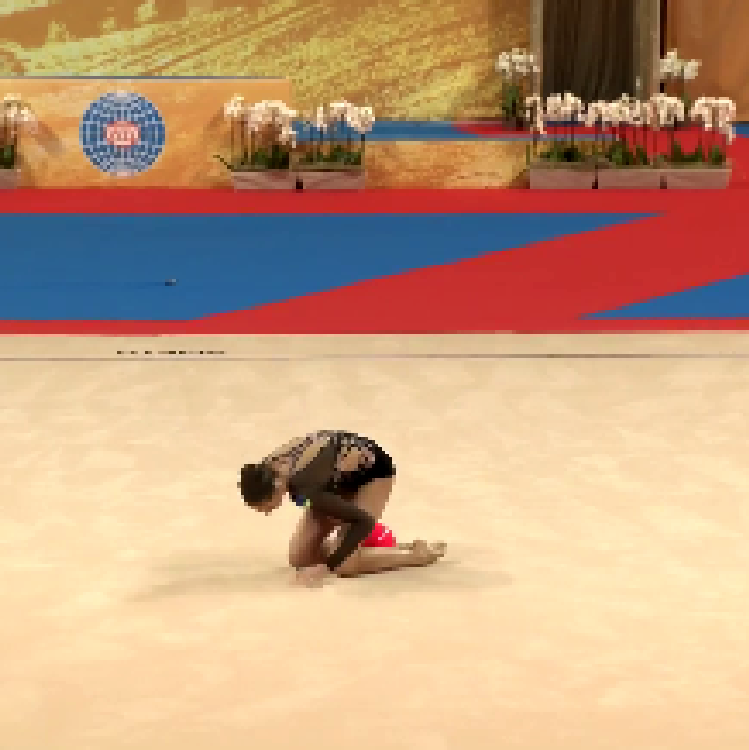}
            \put(3,5){\tiny\color{black}31${\text{-st}}$ frame}
        \end{overpic}%
        \begin{overpic}[width=0.33\linewidth]{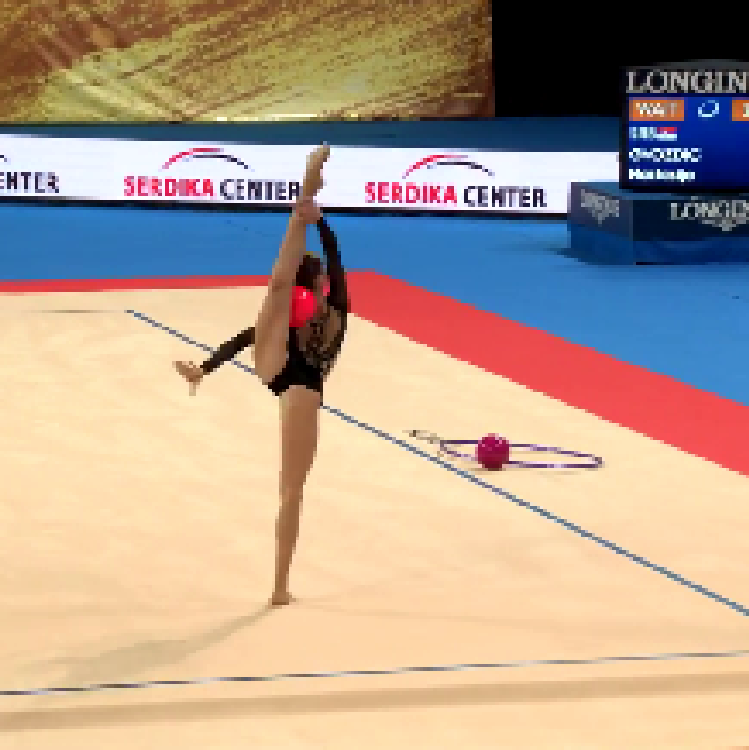}
            \put(3,5){\tiny\color{black}51${\text{-st}}$ frame}
        \end{overpic}                                                                                                                                                               \\[-0.075cm]
        \rotatebox{90}{\scriptsize \color{blue!80} Feature Map} &
        \begin{overpic}[width=0.33\linewidth]{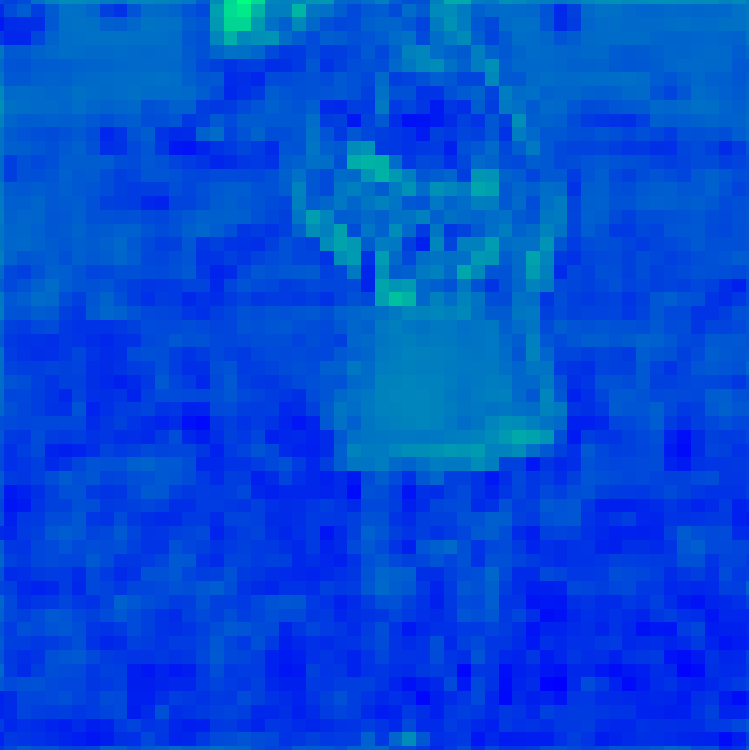}
            \put(3,5){\tiny\color{white}11${\text{-st}}$ frame}
        \end{overpic}%
        \begin{overpic}[width=0.33\linewidth]{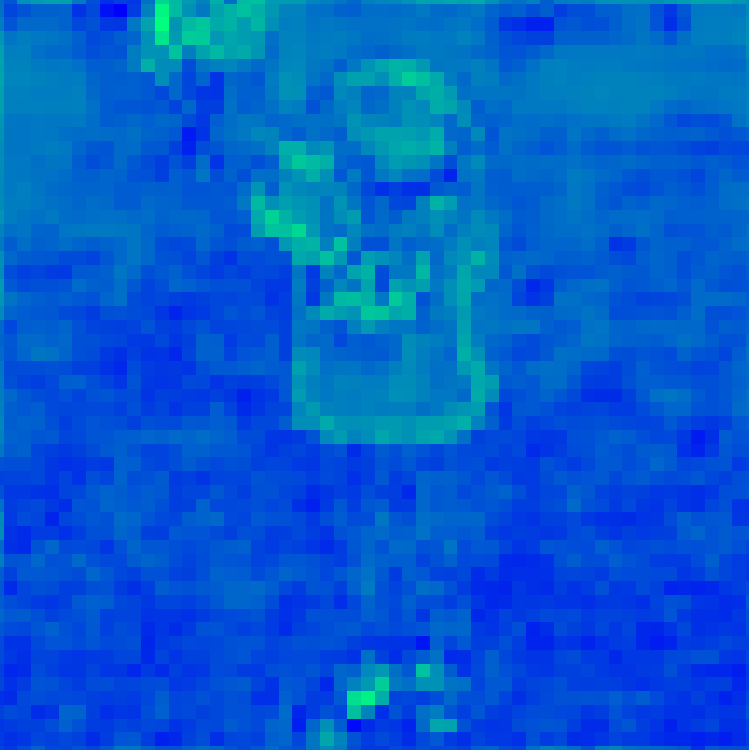}
            \put(3,5){\tiny\color{white}31${\text{-st}}$ frame}
        \end{overpic}%
        \begin{overpic}[width=0.33\linewidth]{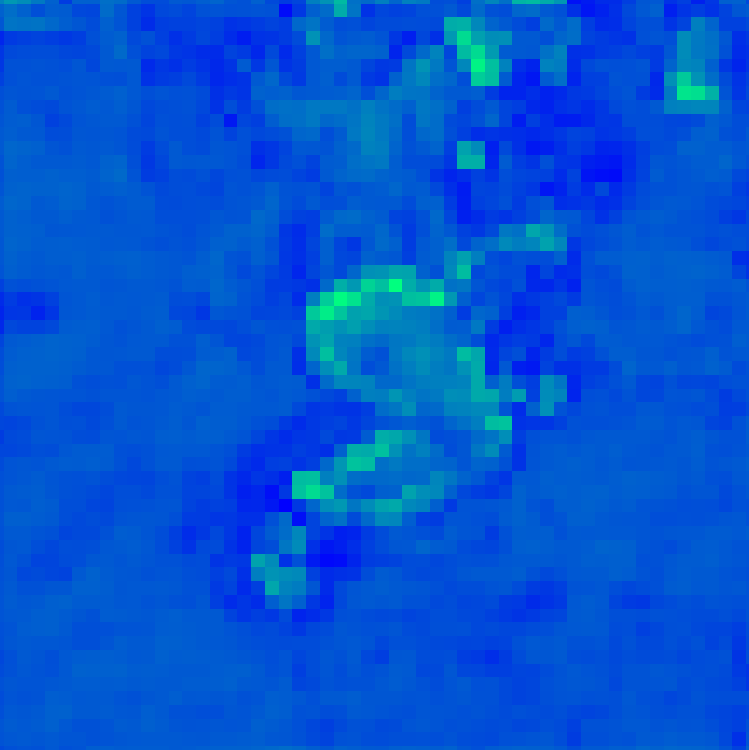}
            \put(3,5){\tiny\color{white}51${\text{-st}}$ frame}
        \end{overpic}%
                                                                &
        \begin{overpic}[width=0.33\linewidth]{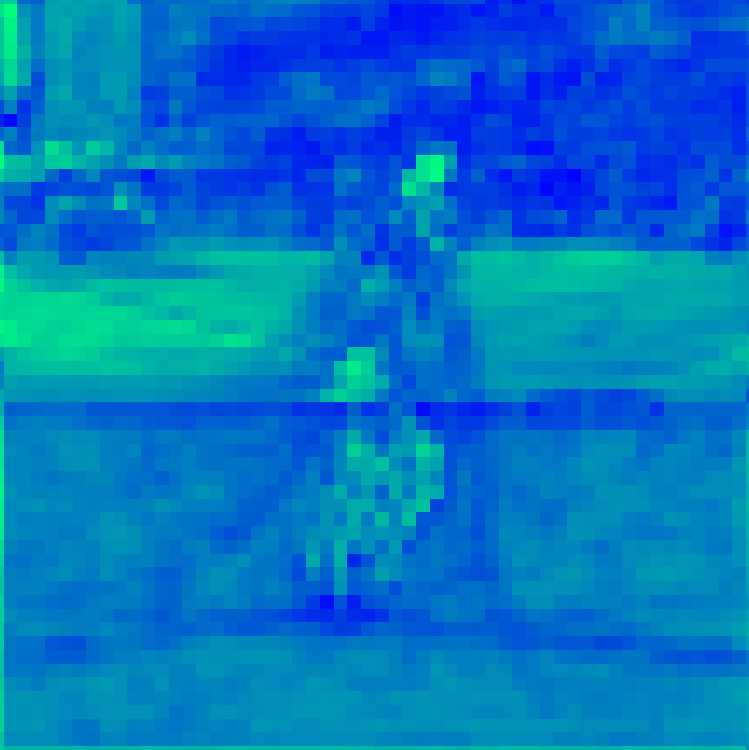}
            \put(3,5){\tiny\color{white}11${\text{-st}}$ frame}
        \end{overpic}%
        \begin{overpic}[width=0.33\linewidth]{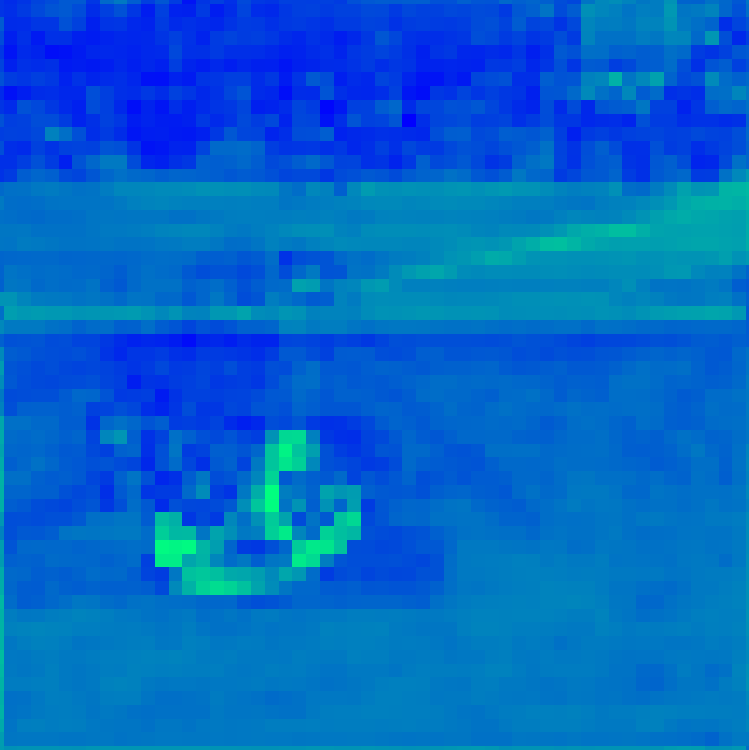}
            \put(3,5){\tiny\color{white}31${\text{-st}}$ frame}
            \put(2,50){\linethickness{0.25mm}\color{yellow}\polygon(0,0)(96,0)(96,40)(0,40)}
            \put(6,75){\tiny\color{white} Score-unrelated}
            \put(45,60){\tiny\color{white} area}
        \end{overpic}%
        \begin{overpic}[width=0.33\linewidth]{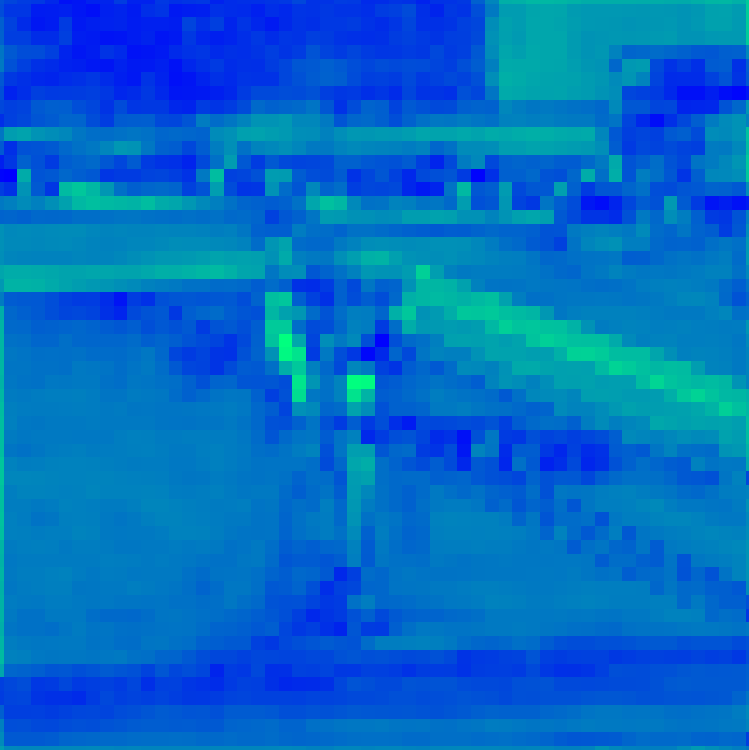}
            \put(2,5){\tiny\color{white}51${\text{-st}}$ frame}
            \put(60,38){\linethickness{0.25mm}\color{yellow}\polygon(0,0)(35,0)(35,60)(0,60)}
        \end{overpic}%
    \end{tabular}
    \begin{overpic}[width=\linewidth,trim=160 155 80 160,clip]{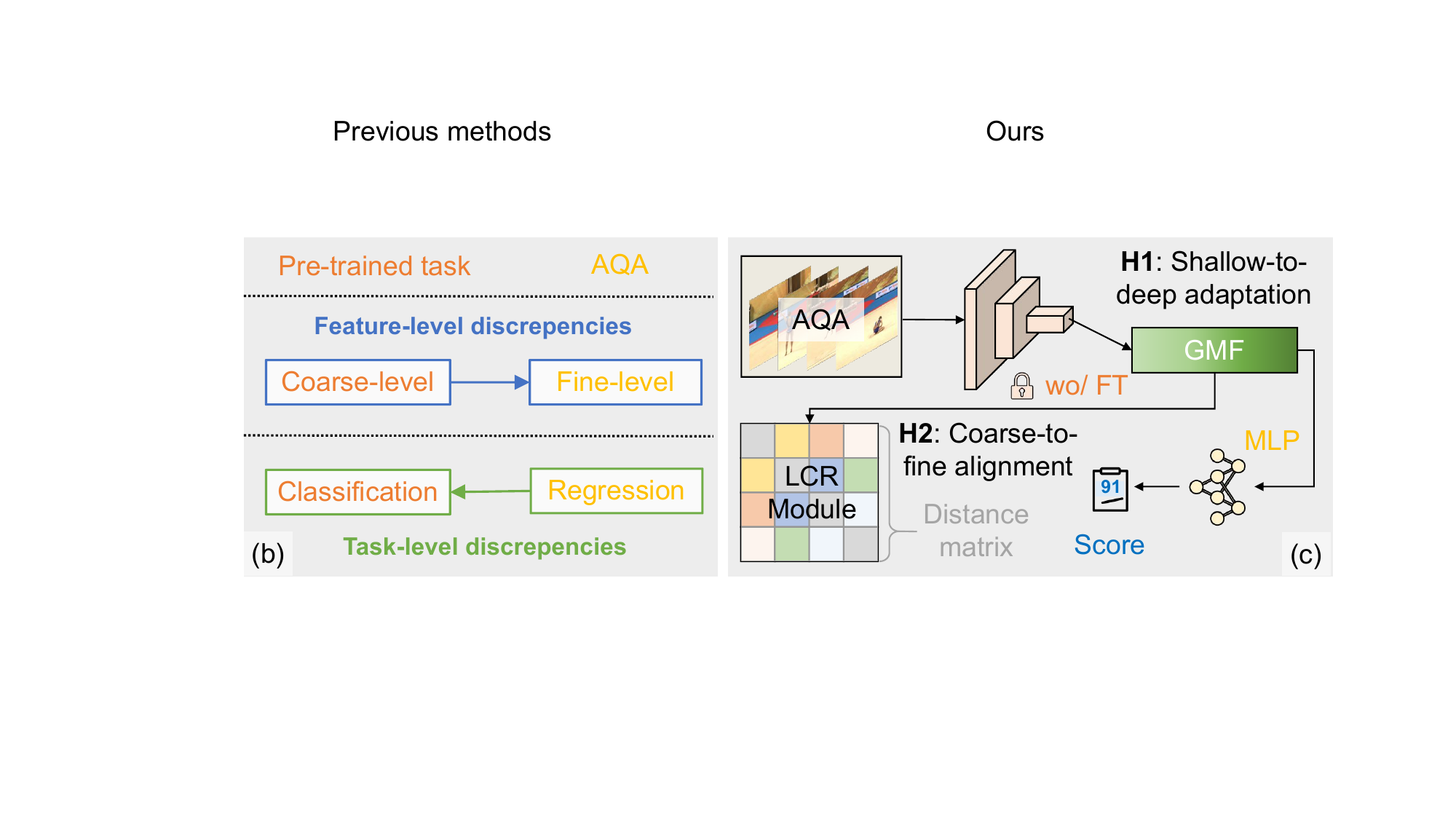}
    \end{overpic}
    \caption{
        Illustrations of our main idea: (a) The pre-trained I3D backbone emphasizes coarse features like guardrails (highlighted in yellow boxes), which may be irrelevant to scoring for AQA, while it can accurately recognize cartwheeling in the action recognition domain. This discrepancy is primarily because the pre-trained task's broader focus on coarse-level features, whereas fine-grained features essential for AQA may not be adequately exploited. (b) We identify two distinct types of domain shift: task-level discrepancies and feature-level discrepancies. (c) Based on two hypotheses, our approach innovates with a shallow-to-deep adaptation using Gap Minimization Flow (GMF), enabling a fast and controllable path to thoroughly minimize the domain gap. Additionally, we introduce a coarse-to-fine alignment mechanism using List-wise Contrastive Regularization (LCR) to enable the model to focus on fine-grained features, essential for AQA, while mitigating domain shift by refining coarse features from the broader pre-trained task.
    }
    \label{fig:idea}{
        \phantomsubcaption\label{fig:idea-a}
        \phantomsubcaption\label{fig:idea-b}
        \phantomsubcaption\label{fig:idea-c}
    }
\end{figure}
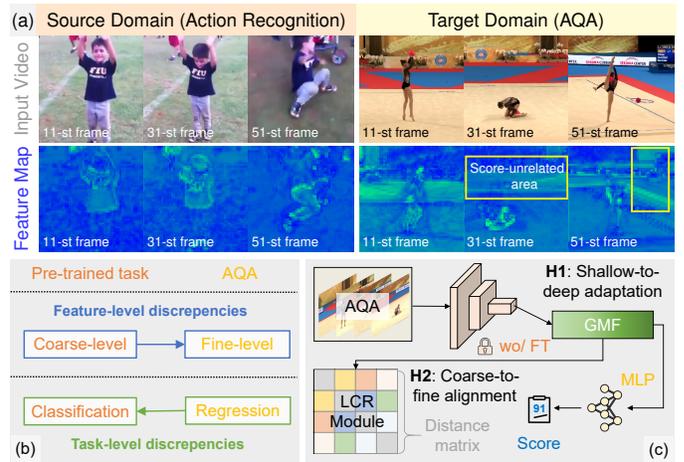

One of the most significant challenges in long-term AQA is the domain shift issue \cite{dadashzadeh2024pecop}, where pre-trained backbones are suboptimal for AQA tasks. This challenge arises from label scarcity and the nature of long video sequences. Firstly, label scarcity contributes to relatively small AQA datasets (e.g., one large-scale long-term AQA dataset for rhythmic gymnastics (balls) with approximately 250 samples). To address this, most AQA methods \cite{parmar2019action,yu2021group,bai2022action} leverage backbones pre-trained on large-scale action recognition datasets (e.g., Kinetics 400 \cite{carreira2017quo} with over 300,000 samples). While this strategy enhances performance on small-scale AQA datasets, its performance is limited by the shift from action recognition to AQA tasks (see \cref{fig:idea-a}).
Secondly, the computational demands of processing these long sequences within long-term AQA, combined with the complexity of resource-intensive backbones \cite{carreira2017quo,liu2022video}, make fine-tuning the backbone impractical with limited computational resources. As a result, existing long-term AQA methods \cite{xu2022likert,zeng2020hybrid} fix the backbone and do not explicitly address the domain shift issue, thereby severely limiting overall performance.
\linelabel{r1:C3.7-b}Indeed, some methods \cite{xu2022likert,zhou2023hierarchical} employing feature aggregation or representation layers, such as Transformers \cite{vaswani2017attention}, can implicitly mitigate domain shift by leveraging score supervision to encourage the model to capture high-level task-relevant patterns. However, these methods remain ineffective for accurate assessment (refer to results in \cref{fig:tsne_ball,fig:tsne}), as they lack explicit adaptation mechanisms to align domain-specific feature distributions.

Long-term AQA aims to evaluate performance based on discriminative features, capturing subtle dynamics over extended periods where domain shift significantly hinders AQA performance.
\linelabel{r1:C3.1}As illustrated in \cref{fig:idea-b}, domain shift arises from both differences in task objectives and variations in data distribution.
Accordingly, we categorize domain shift into task-level discrepancies and feature-level discrepancies. Task-level discrepancies arise from transitioning from classification-based pre-training, where class boundaries are discrete, to regression-based scoring for AQA, where scores vary continuously. Feature-level discrepancies occur due to variations in video capture conditions and application domains, leading pre-trained models to focus on coarse, irrelevant patterns rather than the fine-grained scoring cues necessary for precise AQA.
\linelabel{r1:C3.2}The previous work \cite{zhou2024cofinal} addressed task-level discrepancies by formulating AQA as a coarse-to-fine classification problem (see the alignment arrow ``$\leftarrow$" in the bottom of \cref{fig:idea-b}), but this resulted in precision loss. Our work instead directly tackles feature-level discrepancies by refining pre-trained coarse features to focus on fine-grained cues (see the ``$\rightarrow$" at the top of \cref{fig:idea-b}). This refines pre-trained coarse features to emphasize the fine-grained cues essential for accurate assessment without sacrificing precision.
Experimental results in \cref{tab:sota,tab:sota-fis,tab:sota-logo} demonstrate the superior performance of our approach over addressing only task-level discrepancies, showing the increased importance of minimizing feature-level domain shifts for accurate long-term action assessment.

To this end, we introduce the Progressive Hierarchical Instruction (PHI) framework  (see \cref{fig:idea-c}) to tackle the aforementioned challenges. Built on two major hypotheses regarding shallow-to-deep adaptation and coarse-to-fine alignment, we propose solutions to validate these hypotheses, constituting the core components of PHI.
These approaches collectively reduce the domain gap while prioritizing fine-grained features essential for accurate assessment.

Hypothesis 1: A shallow-to-deep adaptation approach through multi-step control can thoroughly reduce the domain gap between a pre-trained backbone and the AQA task. The rationale behind this lies in the complexity of refining initial features in a single step, whereas the multi-step control facilitates a more progressive and precise gap reduction across shallow to deep layers  (see \cref{fig:one_step}).

To validate Hypothesis 1, we introduce Gap Minimization Flow (GMF) to progressively reduce the domain gap, which is motivated by the recent advances in flow models \cite{liu2022flow,lipman2022flow}.
However, these methods face challenges in constructing training pairs due to the inaccessibility of desired features. To address this, we first integrate a temporally-enhanced attention mechanism to efficiently estimate desired features, capturing crucial long-range dependencies essential for long-term AQA.
This enables GMF to directly and efficiently control the gap reduction, thereby enhancing the model's adaptability.

Hypothesis 2: A coarse-to-fine alignment approach that prioritizes learning representations focusing on fine-grained cues can effectively mitigate domain shift.
The rationale behind this is that pre-trained backbones on large-scale datasets often capture coarse features irrelevant to AQA scoring, whereas AQA depends on fine-grained cues (see \cref{fig:idea-a}).

To validate Hypothesis 2, we present List-wise Contrastive Regularization (LCR) to learn the fine-grained cues essential for AQA, which is motivated by the recent advances in contrastive regression \cite{yu2021group,yao2023contrastive}.
However, these methods are reliant on manual exemplar selection and known exemplar scores during inference, restricting the application scope. To address this, LCR comprehensively compares all pairs of batch data, eliminating the need for manual intervention. In addition, this approach ensures a robust evaluation of action quality, especially in scenarios with limited labeled data.

Experimental results demonstrate significant improvements achieved by our proposed PHI method compared to state-of-the-art methods that do not specifically address domain shift issues. Notably, PHI achieves gains of 5.88\%, 5.65\%, and 24.4\% in correlation on three representative long-term AQA datasets, i.e., RG \cite{zeng2020hybrid}, Fis-V \cite{parmar2017learning}, and LOGO \cite{zhang2023logo}, respectively, compared to shift-unaware methods.
Compared to the task-level solution \cite{zhou2024cofinal}, PHI showcases additional correlation gains and significant precision improvements, demonstrating the superiority of addressing feature-level discrepancies for enhancing long-term AQA performance.
Our source code will be available at \url{https://github.com/ZhouKanglei/PHI_AQA}.
Our main contributions are as follows:
\begin{itemize}
    \item We define domain shifts from both task and feature levels. Our work achieves additional performance gains in long-term AQA by addressing feature-level discrepancies.
    \item We propose a novel gap minimization flow module (GMF) to address the domain gap in a shallow-to-deep manner, facilitating efficient gap reduction control with a fast and straight path.
    \item We design a novel contrastive regularization module (LCR) to mitigate domain shift in a coarse-to-fine manner, enabling robust representation learning for performance improvement.
\end{itemize}

The remainder of this paper is organized as follows: \cref{sec_related_work} reviews the related work in AQA and flow matching methods; \cref{sec_preliminary} briefly describes the preliminaries of rectified flow; \cref{sec_method} details the core components of our proposed framework; \cref{sec_exp} validates and analyzes the effectiveness of our proposed method; \cref{sec_conclusion} concludes the whole paper and outlines the future work.

\section{Related Work} \label{sec_related_work}
In this section, we provide a concise overview of AQA and flow matching, outlining their relevance to our work.

\subsection{Action Quality Assessment (AQA)}
AQA focuses on the quantitative performance of performed actions in various application areas such as sports analysis \cite{zhou2024magr,li2022pairwise,yu2021group,zhou2023hierarchical,yao2023contrastive,han2025finecausal}, medical rehabilitation \cite{zhou2023video,deb2022graph}, and skill assessment \cite{ingwersen2023video,trinh2023self,ding2023sedskill}. Earlier methods depended heavily on hand-crafted features and heuristics, revealing certain limitations. For example, Pirsiavash et al. \cite{pirsiavash2014assessing} leveraged pose features to train a linear SVR model, which was constrained due to the poor pose estimation outcomes \cite{wang2021tsa}. By integrating deep learning, various models have shown improved performance, such as CNNs \cite{zhou2023video,zhou2023hierarchical}, RNNs \cite{xu2019learning,wang2022skeleton}, and Transformers \cite{bai2022action,zhou2024cofinal,zeng2024multimodal,han2025finecausal}.

Existing AQA datasets \cite{zhou2024comprehensive,li2024egoexo} are relatively small, risking over-fitting. To mitigate this, pre-trained backbones are commonly employed. Parmar et al. \cite{parmar2019and} utilized C3D \cite{tran2015learning} to improve AQA performance. Pan et al. \cite{pan2019action} integrated a more robust I3D backbone \cite{carreira2017quo} to further optimize the performance.
Xu et al. \cite{xu2022likert} attempted to incorporate VST \cite{liu2022video}, aiming to derive more powerful features. However, the computational intensity of such 3D backbones presents a notable issue. As every frame may contain essential AQA cues \cite{parmar2017learning}, videos are often segmented into clips for separate processing, which hinders a complete understanding of the action. To this end, Zhou et al. \cite{zhou2023hierarchical} introduced a hierarchical GCN method to eliminate semantic ambiguities. Features extracted are typically aggregated either by LSTM \cite{xu2019learning}, or by more popular average pooling before the score regression.
Instead of methods focusing on single-person AQA datasets, Zhang et al. \cite{zhang2023logo} proposed a group-aware diving dataset.
Furthermore, by leveraging the unique strengths of different data types, such as video, audio, and skeleton data, multi-modal AQA methods \cite{du2024learning,ji2023localization,xia2023skating,zeng2024multimodal,xu2024vision,xu2025dancefix} create a more holistic, accurate, and robust assessment system.
In this work, we primarily focus on RGB-based single-person AQA.

In the context of long-term AQA, the computational intensity of these backbones, combined with the long sequences, poses significant challenges. Since fine-tuning the backbone to minimize the domain shift between the pre-trained broader task and the AQA task is difficult, existing long-term AQA methods \cite{xu2022likert,zeng2020hybrid} often choose to fix the backbone and overlook the domain shift problem, thereby limiting performance.
CoFInAl \cite{zhou2024cofinal} addressed domain shifts by reformulating AQA as coarse and fine classification tasks, potentially leading to a loss in assessment precision. In contrast, our work identifies and tackles the key challenge of domain shift through the lens of feature-level discrepancies. Our method efficiently resolves this issue without the need for fine-tuning the computational backbone and preserving assessment precision.

\subsection{Flow Matching}
Flow Matching (FM) models represent a recent advancement in generative modeling. The term ``flow" refers to a mapping between samples of two distributions, leveraging neural Ordinary Differential Equations (ODEs) to implicitly model the transport plan between simpler and target distributions \cite{tong2023improving}.
These models, such as normalizing flows \cite{albergo2022building,ma2025uncertainty} and continuous normalizing flows \cite{onken2021ot}, have demonstrated impressive capabilities in generating high-quality samples without the need for complex approximate inference techniques \cite{kobyzev2020normalizing}.

Unlike traditional flow-based models, recent approaches propose training algorithms that require solving the ODE explicitly only during inference, overcoming challenges associated with backpropagating through ODEs during training \cite{tong2023improving, lipman2022flow, albergo2022building,liu2022flow,neklyudov2022action}.
FM offers a promising and underexplored avenue for generative modeling, providing a more efficient and effective approach to learning complex distributions.
Different from diffusion models \cite{ho2020denoising,chang2023design,chang2022unifying,wang2025fg,wang2023fg} that utilize Stochastic Differential Equations (SDEs) and are restricted to Gaussian base distributions \cite{song2020score}, FM offers greater flexibility by allowing the choice of base distributions and training with ODEs instead of SDEs, leading to smoother trajectories and improved performance \cite{song2020denoising}.

Our work is inspired by the rectified flow \cite{liu2022flow} that generates a fast and direct flow path but relies on known target samples, which poses significant challenges in addressing domain shift. To overcome this limitation, we introduce a novel approach that does not depend on any known target distributions.
\linelabel{r1:C2.1}We emphasize that our proposed PHI method is not merely an adjustment to rectified flow but a fundamental extension that introduces a novel hierarchical adaptation framework. By leveraging shallow-to-deep adaptation and coarse-to-fine alignment, PHI achieves self-supervised domain adaptation without explicit target distribution samples, making it highly generalizable to a wide range of real-world applications beyond AQA. Notably, PHI is the first attempt at integrating the concept of flow matching in the realm of AQA.

\begin{figure*}
    \centering
    \includegraphics[width=\linewidth,clip,trim=10 125 10 125]{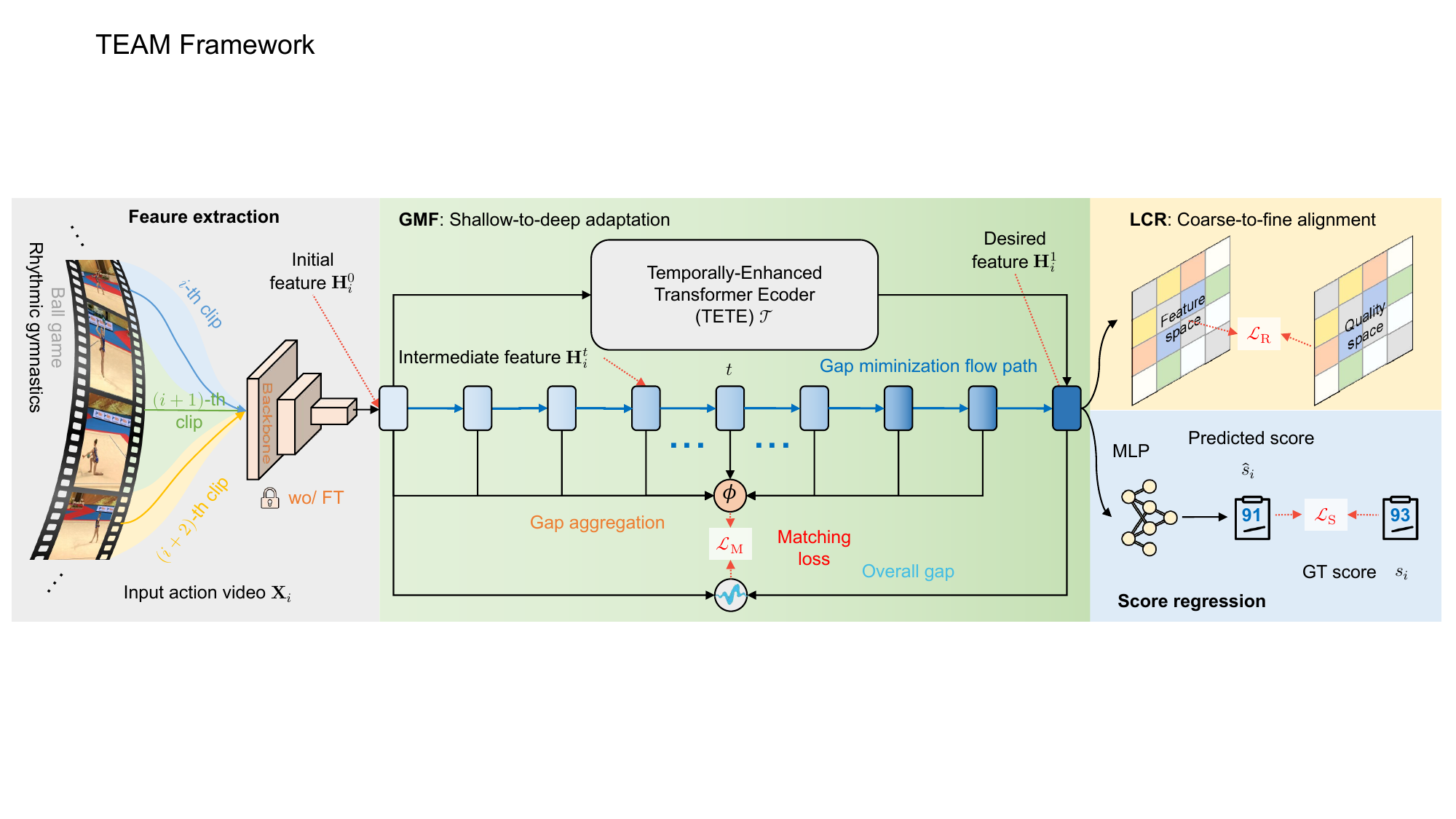}
    \caption{
        Framework of PHI:
        Our PHI framework addresses the domain shift issue through two crucial processes. Firstly, {\color{gcolor}Gap Minimization Flow (GMF)} progressively transforms the initial feature into the desired AQA-specific features, minimizing the domain gap. Secondly, {\color{ycolor}List-wise Contrastive Regularization (LCR)} guides the model towards subtle variations in actions, facilitating the transition from coarse to fine-grained features crucial for AQA. Finally, the refined feature is used to predict the quality score through an MLP.
    }
    \label{fig:framework}
\end{figure*}

\section{Preliminary: Rectified Flow} \label{sec_preliminary}
Rectified Flow (RF) \cite{liu2022flow} offers a straightforward solution for finding a transport map between two observed distributions. It involves learning an ODE, also known as the flow model, aiming to traverse straight paths as much as possible.
Given empirical observations $\bm{x}_0\sim \pi_0, \bm{x}_1\sim \pi_1$, RF finds the transport map implicitly by solving the following ODE:
\begin{equation} \label{eq_ode}
    \mathrm{d} \bm{z}_t = v(\bm{z}_t, t) \mathrm{d} t, \quad t\in[0,1],
\end{equation}
where $v\colon \mathbb{R}^d\to \mathbb{R}^d$ represents the drift force that converts $\bm{z}_0$ from distribution $\pi_0$ to $\bm{z}_1$ following distribution $\pi_1$.

RF operates by aligning the ODE with the linear interpolation of points from $\pi_0$ and $\pi_1$.
Given $\bm{x}_0$ and $\bm{x}_1$, the linear interpolation of $\bm{x}_0$ and $\bm{x}_1$ is $\bm{x}_t = t \bm{x}_1 + (1-t) \bm{x}_0$.
Observe $\bm{x}_t$ follows a trivial ODE that already transfers $\pi_0$ to $\pi_1$, i.e., $\mathrm{d}  \bm{x}_t =(\bm{x}_1-\bm{x}_0)\mathrm{d} t$, where $\bm{x}_t$ moves following the line direction $(\bm{x}_1-\bm{x}_0)$ with a constant speed. However, this ODE does not solve the problem: it cannot be simulated causally, because the update $\bm{x}_t$  depends on the final state $\bm{x}_1$, which is not supposed to be known at time $t<1$.

RF casualizes the interpolation process by projecting it onto the space of causally simulatable ODEs in \cref{eq_ode}.
The drift force $v$ is set to drive the flow to follow the direction $(\bm{x}_1-\bm{x}_0)$ as much as possible. A natural way to the $\ell_2$ projection on the velocity field is finding $v$ by solving a simple least squares regression problem:
\begin{equation}
    \min_{v}~
    \int_0^1 \mathbb{E} \left[\| {( \bm{x}_1 - \bm{x}_0) - v (\bm{x}_t,~ t)}\|^2
    \right] \mathrm{d} t.
    \label{equ:mainf}
\end{equation}
By fitting $v$ with the direction $(\bm{x}_1-\bm{x}_0)$, RF casualizes the paths of linear interpolation $\bm{x}_t$, yielding an ODE flow that can be simulated without seeing the future.
We can parameterize $v$ with a neural network and solve \cref{equ:mainf} with any stochastic optimizer.

While RF aims to find a transport map between two observed distributions by projecting the interpolation path onto a space of causally simulatable ODEs, our proposed Gap Minimization Flow module extends this concept to progressively minimize the domain gap between the pre-trained backbone and the target AQA task. Unlike the ODE-based formulation in RF, our approach leverages a temporally-enhanced attention mechanism to efficiently estimate the desired features, allowing for a more targeted and effective domain gap reduction.

\section{Methodology: PHI} \label{sec_method}
This section first introduces the PHI framework, followed by a detailed explanation of its core components.

\paragraph{Problem Definition}
Long-term AQA involves assessing the quality or proficiency of actions through extended video recordings, considering temporal factors such as consistency, progression, and the detailed evolution of the comprehensive performance over time, rather than just snapshots.
This presents significant challenges in addressing the domain shift issue, modeling long-range temporal dependencies, and capturing fine-grained dynamics.

\paragraph{Framework Overview}
Our PHI method (see \cref{fig:framework}) addresses the aforementioned challenges through a holistic two-stage methodology comprising shallow-to-deep adaptation and coarse-to-fine alignment.
Given an action video $\mathbf{X}_i\in \mathbb{R}^{T\times W \times H \times 3}$, representing $T$ frames of resolution $W\times H$ and $3$ color channels, the initial feature $\mathbf{H}_i^0\in \mathbb{R}^{M \times D}$ is extracted using a pre-trained backbone, where $M$ denotes the number of clips divided from the video.
In the shallow-to-deep adaptation stage (see \cref{sec_s2d}), Gap Minimization Flow (GMF) with temporally-enhanced attention transforms the initial feature $\mathbf{H}_i^0$ into the desired feature representation $\mathbf{H}_i^1\in \mathbb{R}^{M \times D}$.
The loss function $\mathcal{L}_{\mathrm{M}}$ (see \cref{eq_flow}) facilitates efficient adjustment of the initial feature to achieve the desired representation that better aligns with AQA.
In the coarse-to-fine alignment stage (see \cref{sec_c2f}), List-wise Contrastive Regularization (LCR) is employed to regularize the feature space.
The regularization loss $\mathcal{L}_{\mathrm{R}}$ (see \cref{eqcr}) encourages the adaptation process to prioritize fine-grained features essential for AQA.
Ultimately, the desired feature representation $\mathbf{H}_i^1$ is fed into an MLP head network to predict the quality score $\hat{s}_i$, which is supervised by the MSE loss $\mathcal{L}_{\mathrm{S}}=\frac{1}{2}\sum_i(s_i - \hat{s}_i)^2$. Overall, the total loss can be represented as:
\begin{equation}
    \mathcal{L} = \mathcal{L}_{\mathrm{S}} + \lambda_{\mathrm{M}} \mathcal{L}_{\mathrm{M}} + \lambda_{\mathrm{R}} \mathcal{L}_{\mathrm{R}},
\end{equation}
where $\lambda_{\mathrm{M}}$ and $\lambda_{\mathrm{R}}$ are loss weights.
During inference, the initial feature only undergoes the flow path to regress the score.

\subsection{GMF: Shallow-to-Deep Adaptation}
\label{sec_s2d}
\subsubsection{Design Idea}
\begin{hyp} \label{h1}
    A shallow-to-deep adaptation approach through multi-step control can thoroughly reduce the domain gap between a pre-trained backbone and the AQA task.
\end{hyp}

\paragraph{Justification}
The challenge of mitigating domain shift arises from the complexity of transferring complex data between initial and desired features in a single step, often compromising reliability and precision. A multi-step control mechanism involves breaking down the distribution gap into multiple sub-parts and sequentially matching these distributions. This approach enables gradual and accurate feature transformation, facilitating a thorough reduction of the domain gap across shallow to deep layers.

\paragraph{Implementation}
The core idea of GMF is to utilize the concept of RF \cite{liu2022flow} to gradually reduce the domain gap, which relies on paired samples to train the flow path. However, in our context, we lack the corresponding desired feature for the initial feature, posing a significant challenge in minimizing the domain gap. To address this, we first propose to estimate the desired feature and then progressively minimize the gap. Fortunately, the semantics of the desired feature are known and represented by the score. Thus, leveraging score loss $\mathcal{L}_{\mathrm{S}}$ allows for the coarse estimation of the desired feature.
Our shallow-to-deep adaptation consists of two steps: desired feature estimation and domain gap minimization.

\subsubsection{Desired Feature Estimation} \label{sec_s2d-att}
\begin{figure}
    \centering
    \includegraphics[width=\linewidth,trim=90 40 82 100,clip]{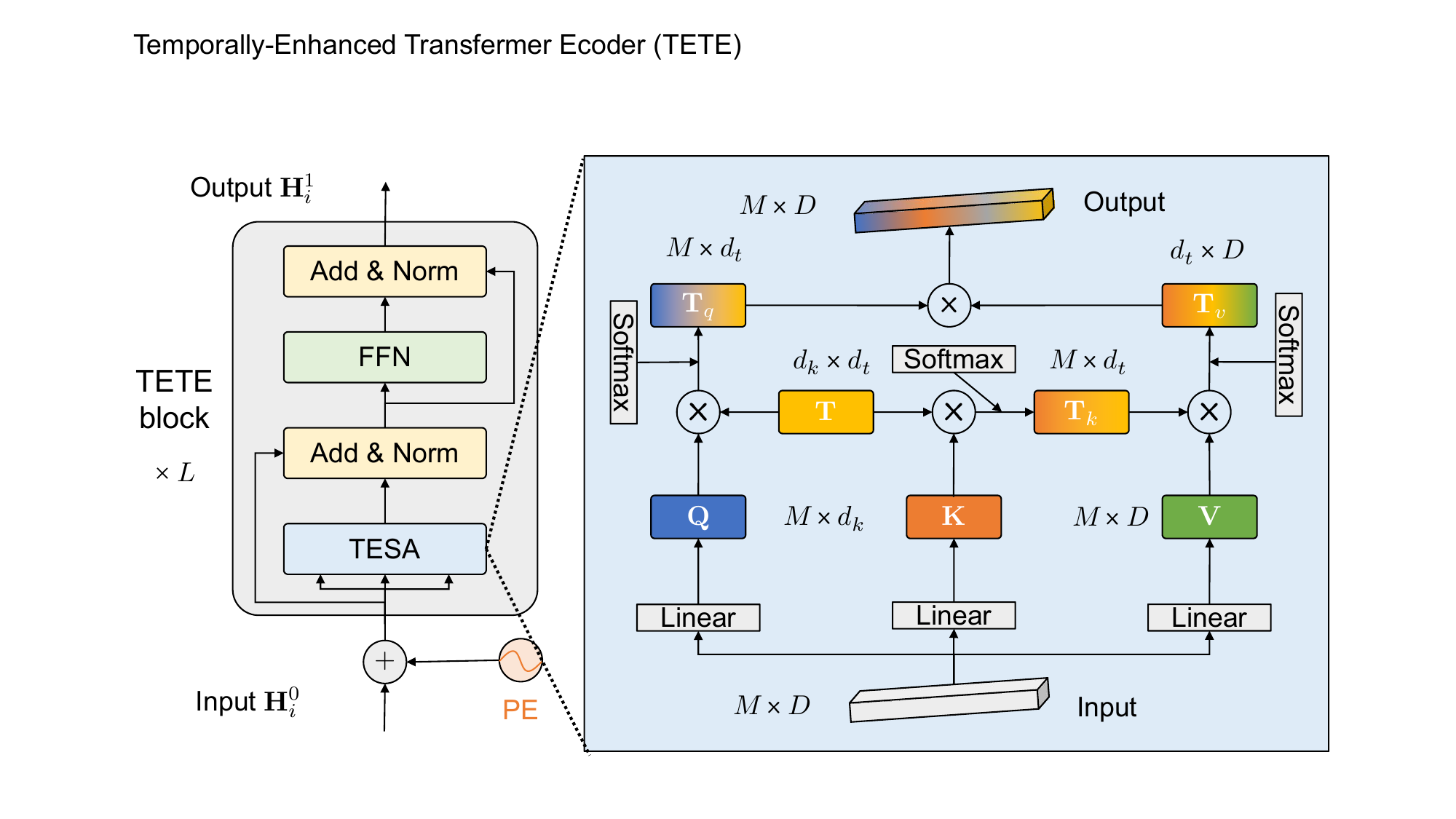}
    \caption{Illustration of Temporally-Enhanced Transformer Encoder (TETE): Highlighting the attention of Temporally-Enhanced Self-Attention (TESA), we employ the low-rank matrix to reduce the complexity from $\mathcal{O}(M^2D)$ to $\mathcal{O}(Md_{t}D)$, ensuring efficient modeling of long-term dependencies.}
    \label{fig:tete}
\end{figure}

In long-term AQA, effectively modeling long-range dependencies in sequences is crucial.
\linelabel{r1:C3.2-b}The previous work \cite{zhou2024cofinal} employs a simple transformation matrix to aggregate temporal dependencies, which, while effective, cannot capture long-range temporal relationships.
While self-attention can model the long-range dependencies, it entails substantial computational overhead. To address this, we propose Temporally-Enhanced Transformer Encoder (TETE, see \cref{fig:tete}) using low-rank decomposition to optimize computation within attention. TETE can efficiently model long-range dependencies in long-term AQA by reducing computational complexity in attention mechanisms and enabling accurate estimation of desired features.

\paragraph{Vanilla Self-Attention}
Considering the initial feature $\mathbf{H}_i^0 \in \mathbb{R}^{M\times D}$ of $i$-th action video, the vanilla self-attention \cite{vaswani2017attention} can be represented as:
\begin{equation} \label{eq_self}
    \mathrm{Attention}(\mathbf{Q},\mathbf{K},\mathbf{V}) = \mathrm{Softmax}\left(\frac{\mathbf{Q}\mathbf{K}^\top}{\sqrt{d_k}}\right)\mathbf{V},
\end{equation}
where $\mathbf{Q} \in \mathbb{R}^{M \times d_k},\mathbf{K} \in \mathbb{R}^{M \times d_k},\mathbf{V} \in \mathbb{R}^{M \times D}$ are the linear embeddings of the input $\mathbf{H}_i^0$. The computation complexity of \cref{eq_self} is $\mathcal{O}(M^2 D)$, where the clip number $M$ is typically large for long-term AQA. Thus, computation increases sharply with longer sequences in real-world scenarios.

\paragraph{Temporally-Enhanced Self-Attention (TESA)} Our proposed TESA module addresses the computational challenges of long sequences by learning a low-rank matrix $\mathbf{T} \in \mathbb{R}^{d_t \times d_k}$,
\linelabel{r1:C1.2}\linelabel{r1:C1.1-b}where $d_t$ is significantly smaller than the feature dimension $D$, ensuring computational efficiency while preserving temporal dependencies.
\linelabel{r1:C1.1}First, TESA applies $\mathbf{T}$ only to the query $\mathbf{Q}$ and the key $\mathbf{K}$, as they directly determine the attention weights. This enhances the ability of the model to capture long-term dependencies while avoiding redundant transformations on the value $\mathbf{V}$, which does not influence attention weight computation. The transformed queries and keys are computed as:
\begin{align}
    \mathbf{T}_q = \mathrm{Softmax}\left(\mathbf{Q} \mathbf{T}^\top \right) \in \mathbb{R}^{M \times d_t}, \\
    \mathbf{T}_k = \mathrm{Softmax}(\mathbf{K} \mathbf{T}^\top )\in \mathbb{R}^{M \times d_t}.
\end{align}
Next, we use $\mathbf{T}_k$ to query the value $\mathbf{V}$, yielding:
\begin{align}
    \mathbf{T}_v = \mathrm{Softmax}(\mathbf{T}_k \mathbf{V}) \in \mathbb{R}^{d_t \times D}.
\end{align}
Finally, our attention mechanism can be represented as:
\begin{equation} \label{eq_our_self}
    \mathrm{Attention}\left(\mathbf{Q},\mathbf{K},\mathbf{V},\mathbf{T}\right) = \mathbf{T}_q  \mathbf{T}_v  \in \mathbb{R}^{M \times D}.
\end{equation}
This results in a decent complexity reduction to $\mathcal{O}(Md_tD)$, effectively improving the training efficiency and mitigating overfitting, thus leading to performance improvement (refer to results in \cref{tab:ablation}).

Finally, the desired feature can be estimated as:
\begin{equation}
    \mathbf{H}_i^1 = \mathcal{T}\left(\mathbf{H}_i^0\right),
\end{equation}
where $\mathcal{T}(\cdot)$ represents the TETE module.

\subsubsection{Domain Gap Minimization}
Inspired by the concept of RF \cite{liu2022flow}, our work aims to learn a fast and controllable path to efficiently mitigate domain shift by progressively reducing the domain gap between initial and desired features.

Suppose the initial feature $\mathbf{H}_i^0$ and the desired feature $\mathbf{H}_i^1$ follow two different empirical distributions, and we sample $P$ intermediate steps. The target representation of the $j$-th step can be defined as the linear interpolation  $\mathbf{H}_i^{j/P} = \frac{j}{P} \mathbf{H}_i^0 + (1-\frac{j}{P}) \mathbf{H}_i^1$. During training, our objective is to predict the next step using the previous representation and the step size through a neural network $\phi(\cdot,\cdot)$, which models the domain gap of the corresponding step.
In practice, our experiments demonstrate that a simple MLP is sufficient for this task.
The gap at the $j$-th step can be represented as:
\begin{equation}
    \bm{g}_j = \phi\left(\hat{\mathbf{H}}_i^{(j-1)/P}, \frac{1}{P}\right), \quad ~ j \geq 1,
\end{equation}
where $\hat{\mathbf{H}}_i^{(j-1)/P}$ denotes the predicted feature of the previous step, calculated as $\bm{g}_{j-1}+\hat{\mathbf{H}}_i^{(j-2)/P}$ if $j \geq 2$, otherwise $\mathbf{H}_i^0$.
The network $\phi$ generates a driving force that guides the flow in the direction of the overall flow, and we optimize $\phi$ by:
\begin{align} \label{eq_flow}
    \mathcal{L}_{\mathrm{M}}        & = \frac{1}{B}\sum_{i=0}^{B-1}\left(
    \mathcal{L}_{\mathrm{M-global}}
    +
    \mathcal{L}_{\mathrm{M-local}}
    \right)
    ,                                                                                                                                \\
    \mathcal{L}_{\mathrm{M-global}} & = \left\| \left(\mathbf{H}_i^1 - \mathbf{H}_i^0\right) - \sum_{j=1}^{P} \bm{g}_{j} \right\|^2, \\
    \mathcal{L}_{\mathrm{M-local}}  & = \frac{1}{P}\sum_{j=1}^P \left\| \mathbf{H}_i^{j/P} - \hat{\mathbf{H}}_i^{j/P} \right\|^2,
\end{align}
where $B$ denotes the batch size, and $\left(\mathbf{H}_i^1 - \mathbf{H}_i^0\right)$ represents the overall flow direction. The first term $\mathcal{L}_{\mathrm{M-global}}$ represents the global flow constraint, while the second term $\mathcal{L}_{\mathrm{M-local}}$ denotes the local flow constraint. These two terms are collectively effective in addressing different aspects of the flow constraints.

\subsubsection{Benefits of GMF}
\linelabel{r1:C2.1-b}GMF introduces a fundamentally new perspective on feature transformation in flow-based adaptation, offering several key advantages over existing methods. Unlike traditional flow methods \cite{lipman2022flow,liu2022flow}, which require explicit knowledge of the target distribution, GMF eliminates this dependency by leveraging an adaptive hierarchical transformation strategy. This enables GMF to handle domain-shifted tasks where the target distribution is either unknown or difficult to obtain, significantly extending the applicability of flow-based learning. Instead of directly aligning source and target distributions, GMF sequentially matches intermediate representations through a multi-stage adaptation process. This hierarchical alignment ensures a smooth and stable transition while preventing abrupt transformations. Unlike stacked or recurrent architectures, GMF dynamically regulates the adaptation trajectory, reducing computational overhead and improving convergence efficiency. Additionally, its implicit knowledge distillation mechanism allows for lightweight inference without compromising performance. These innovations make GMF a scalable and efficient solution for domain-adaptive learning in video-based tasks, surpassing conventional flow-based approaches.

\subsection{LCR: Coarse-to-Fine Alignment}
\label{sec_c2f}
\subsubsection{Design Idea}

\begin{hyp} \label{h2}
    A coarse-to-fine alignment approach that prioritizes learning representations focusing on fine-grained cues can effectively mitigate domain shift in long-term AQA.
\end{hyp}

\paragraph{Justification}
As shown in \cref{fig:idea-a}, the pre-trained models on broader tasks often emphasize coarse-level features that may not directly align with the requirements of AQA.
Conversely, AQA tasks rely heavily on fine-grained cues \cite{gedamu2023fine,xu2022finediving} for accurate assessment. Therefore, a method that prioritizes learning these fine-grained representations is likely to be more effective in addressing the domain shift issue.

\begin{figure}
    \centering
    \includegraphics[width=\linewidth,trim=80 100 85 100,clip]{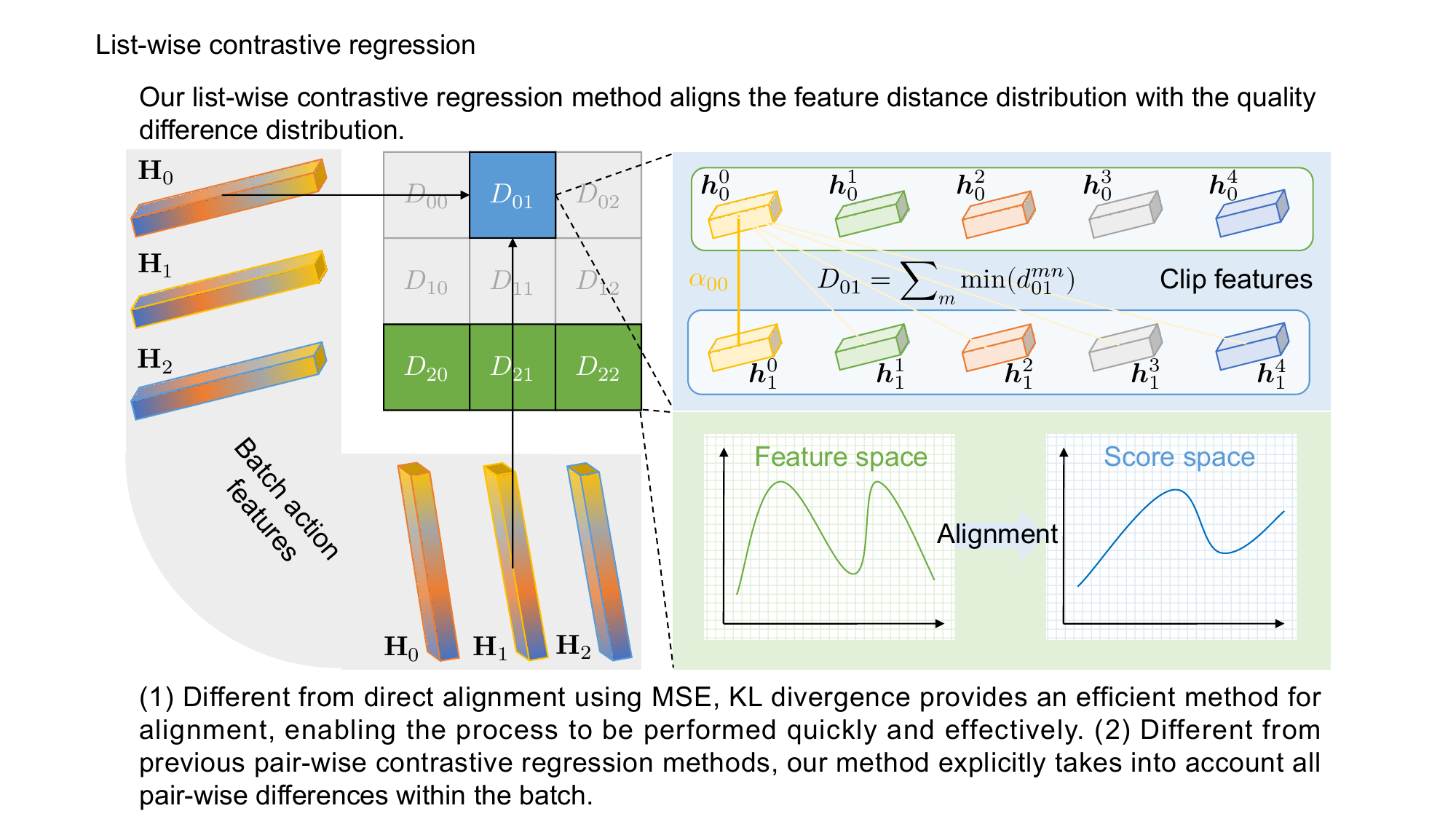}
    \caption{Illustration of List-wise Contrastive Regularization (LCR): LCR aligns the distance distribution in the feature space with that of the quality score space in a list-wise manner. This ensures comprehensive comparison and alignment across the entire batch of data, leading to robust performance.}
    \label{fig:lcr}
\end{figure}

\paragraph{Implementation}
\linelabel{r1:C3.2-c}Notably, the interpretation of coarse-to-fine alignment differs from CoFInAl \cite{zhou2024cofinal}. In CoFInAl, coarse-to-fine alignment refers to structuring AQA as a two-stage classification task, which helps address task-level differences but may introduce precision loss. In contrast, PHI adopts a feature refinement approach that progressively enhances pre-trained coarse features to focus on fine-grained scoring cues, ensuring improved AQA performance.
In response, our novel LCR module (see \cref{fig:lcr}) employs a coarse-to-fine alignment strategy to further address the domain shift, enhancing the model's ability to capture fine-grained features essential for AQA.
By comparing the differences between actions, contrastive regression aids in identifying subtle variations or abnormalities that may not be apparent when assessing actions in isolation.
Given a batch of data, LCR involves computing a distance matrix where each row encodes the relationships of an action with all other actions in the batch. By aligning the distribution of each row with its ground truth quality score distribution, the model can effectively learn subtle variations.

\subsubsection{Action Distance Computation}
\linelabel{r1:C3.7-c}For each action in the batch, we need to calculate its pairwise distance with all other actions. However, directly measuring the distance between two actions is inappropriate because their clips may not be temporally or semantically aligned. For instance, clips with similar semantics might appear at different indices in each action sequence. Hence, we need to establish correspondence between clips before computing the action-level distance.
(1) To align clips in two actions, for each clip in the first action, we compute its distance to all clips in the second action and select the closest one as its paired clip. Specifically, the pairing is determined based on the minimum $\ell_2$ distance.
(2) Then, we sum all the clip pair distances to obtain the final distance between the two actions. This process can be represented as:
\begin{equation}
    D_{ij} = \sum_{m=0}^{M-1} \left( \min_{n\in [0, M-1]}(d_{ij}^{mn}) \right),~ d_{ij}^{mn} = \| \bm{h}_i^m - \bm{h}_j^n\|^2,
\end{equation}
where $d_{ij}^{mn}$ denotes the distance between the $m$-th clip of the $i$-th action video and the $n$-th clip of the $j$-th action video using the $\ell_2$ distance.

It is interesting to note the implicit temporal order relationship between the paired clip distance calculation. In practice, when considering two clips $\bm{h}_i^{m_1}$ and $\bm{h}_i^{m_2}~ (m_1 \leq m_2)$ from the same video, where $m_1$ and $m_2$ represent different time steps, and their paired clips $\bm{h}_j^{n_1}$ and $\bm{h}_j^{n_2}$, it is observed that upon convergence, $n_1$ tends to be less than or equal to $n_2$. This observation aligns with the inherent temporal ordering within action sequences, where paired clips that are closer in time tend to have smaller indices than those further apart. Therefore, when we added a loss item to constrain $n_1 \leq n_2$, the performance did not change.

\subsubsection{Distance Matrix Alignment} Aligning the distribution of each row in the distance matrix with the ground truth quality score distribution involves two main steps. (1) We calculate the ground truth quality score distance using $\mathbf{S} = |\bm{s} - \bm{s}^\top|$, where $\bm{s}\in {\mathbb{R}^{B\times 1}}$ denotes the quality score of the batch data ($B$ is the batch size). (2) Optimizing the alignment process can be challenging, as traditional metrics like MSE may not accurately capture the underlying structure of the data (see results in \cref{tab:ablation}). To address this issue, we propose incorporating Kullback-Leibler (KL) divergence \cite{raiber2017kullback} to enhance the alignment process and better capture the complex relationships within the data distributions.
This process can be optimized by:
\begin{equation} \label{eqcr}
    \begin{aligned}
        \mathcal{L}_{\mathrm{R}} & = \sum_{i=0}^{B-1} \left( \mathrm{KL}(\bm{S}_i \| \bm{D}_i) + \mathrm{KL}(\bm{D}_i \| \bm{S}_i)) \right).
    \end{aligned}
\end{equation}

\subsubsection{Benefits of LCR}
\linelabel{r1:C2.1-c}LCR introduces a novel hierarchical alignment strategy that significantly enhances AQA performance by capturing subtle relationships across batch samples. Unlike traditional pair-wise contrastive methods \cite{ke2024two,yu2021group,gedamu2023fine}, LCR employs batch-wise contrastive learning to explicitly consider all pair-wise differences, enabling a more effective capture of fine-grained variations. The loss function $\mathcal{L}_{\mathrm{R}}$ leverages KL divergence for efficient alignment between distributions
$\bm{D}_i$ and $\bm{S}_i$, ensuring faster convergence and better generalization compared to direct alignment methods like MSE. Additionally, implicit low-rank regularization enforces temporal coherence and robustness, preventing overfitting to spurious correlations. The batch-wise learning and low-rank constraints collectively establish LCR as a theoretically unique and generalizable framework for domain adaptation, applicable not only to AQA but also to tasks like rehabilitation analysis, sports motion scoring, and action recognition. By addressing the limitations of prior methods, LCR represents a significant advancement in feature alignment for domain-shifted tasks. For instance, it can be adaptively applied to address dynamic data distribution challenges \cite{zhou2025adaptive,zhou2024magr}.

\begin{table*}
    \centering
    \rowcolors{19}{gray!12}{gray!2}
    \setlength{\tabcolsep}{0.5em}
    \caption{
        Results of SRCC ($\uparrow$) and R-$\ell_2$ ($\downarrow$) on the RG dataset.
        The best results are highlighted in \textbf{bold}, while the second-best results are \underline{underlined}.
        The symbol ``$\star$" indicates our reimplementation based on the official code.
        The average SRCC is computed using the Fisher-z value.
        ``--" denotes that the method does not report this metric.
        ``+" indicates the use of additional features/modalities.
    }
    \resizebox{\linewidth}{!}{
        \begin{tabular}{rrcccrrrrrrrrrrrrrrrcccccccccccccc}
            \toprule
            \multirow{2}{*}[-1ex]{\bf Method}                 &
            \multirow{2}{*}[-1ex]{\bf Publisher}              &
            \multirow{2}{*}[-1ex]{\bf Backbone}               &
            \multirow{2}{*}[-1ex]{\makecell[c]{\bf Shift                                                                                                                                                                                                                                                                                                                                                                     \\ \bf Aware}}                 &
            \multirow{2}{*}[-1ex]{\makecell[c]{\bf Modality}} &
            \multicolumn{2}{c}{\bf Ball}                      & \multicolumn{2}{c}{\bf Clubs} & \multicolumn{2}{c}{\bf Hoop} & \multicolumn{2}{c}{\bf Ribbon} &
            \multicolumn{2}{c}{\bf Average}                                                                                                                                                                                                                                                                                                                                                                                  \\
            \cmidrule(lr){6-7}\cmidrule(lr){8-9}\cmidrule(lr){10-11}\cmidrule(lr){12-13}\cmidrule(lr){14-15}
                                                              &                               &                              &                                &                   & \textbf{SRCC}     & \textbf{R-$\bm{\ell_2}$} & \textbf{SRCC}     & \textbf{R-$\bm{\ell_2}$} & \textbf{SRCC}     & \textbf{R-$\bm{\ell_2}$} & \textbf{SRCC}     & \textbf{R-$\bm{\ell_2}$} & \textbf{SRCC}     & \textbf{R-$\bm{\ell_2}$} \\
            \midrule
            C3D+SVR \cite{parmar2017learning}                 & TPAMI'17                      & C3D                          & \ding{53}                      & RGB               & 0.357             & --                       & 0.551             & --                       & 0.495             & --                       & 0.516             & --                       & 0.483             & --                       \\
            \hdashline
            MS-LSTM \cite{xu2019learning}                     & TCSVT'19                      & I3D                          & \ding{53}                      & RGB               & 0.515             & --                       & 0.621             & --                       & 0.540             & --                       & 0.522             & --                       & 0.551             & --                       \\
            ACTION-NET \cite{zeng2020hybrid}                  & ACM MM'20                     & I3D+                         & \ding{53}                      & RGB               & 0.528             & --                       & 0.652             & --                       & 0.708             & --                       & 0.578             & --                       & 0.623             & --                       \\
            GDLT \cite{xu2022likert}                          & CVPR'22                       & I3D                          & \ding{53}                      & RGB
                                                              & 0.526                         & \underline{2.943}            & {0.710}                        & \textbf{2.557}    & {0.729}           & {8.149}                  & {0.704}           & \underline{3.485}        & {0.674}           & \underline{4.284}                                                                                                      \\
            HGCN$^\star$ \cite{zhou2023hierarchical}          & TCSVT'23                      & I3D                          & \ding{53}                      & RGB
                                                              & {0.534}                       & 6.748                        & 0.609                          & 16.142            & 0.706             & 9.270                    & 0.621             & 9.934                    & 0.621             & 10.524                                                                                                                 \\
            VATP-Net \cite{gedamu2024visual}                  & TCSVT'24                      & I3D                          & \ding{53}                      & RGB+              & 0.580             & --                       & 0.720             & --                       & 0.739             & --                       & 0.724             & --                       & 0.709             & --                       \\
            CoFInAl \cite{zhou2024cofinal}                    & IJCAI'24                      & I3D                          & \ding{51}                      & RGB
                                                              & \textbf{0.625}                & \textbf{2.647}               & \underline{0.719}              & \underline{3.093} & \textbf{0.734}    & \textbf{3.892}           & \textbf{0.757}    & \textbf{2.607}           & \textbf{0.712}    & \textbf{3.060}                                                                                                         \\
            PHI (Ours)                                        & --                            & I3D                          & \ding{51}                      & RGB
                                                              & \underline{0.598}             & {3.471}                      & \textbf{0.732}                 & {3.139}           & \underline{0.731} & \underline{5.376}        & \underline{0.754} & {5.674}                  & \underline{0.708} & {4.415}                                                                                                                \\
            \hdashline
            MS-LSTM \cite{xu2019learning}                     & TCSVT'19                      & VST                          & \ding{53}                      & RGB               & 0.621             & --                       & 0.661             & --                       & 0.670             & --                       & 0.695             & --                       & 0.663             & --                       \\
            ACTION-NET \cite{zeng2020hybrid}                  & ACM MM'20                     & VST+                         & \ding{53}                      & RGB               & 0.684             & --                       & 0.737             & --                       & 0.733             & --                       & {0.754}           & --                       & 0.728             & --                       \\
            GDLT \cite{xu2022likert}                          & CVPR'22                       & VST                          & \ding{53}                      & RGB               & {0.746}           & {2.833}                  & {0.802}           & \underline{2.179}        & {0.765}           & \textbf{2.012}           & 0.741             & \underline{2.579}        & {0.765}           & \underline{2.401}        \\
            HGCN$^\star$ \cite{zhou2023hierarchical}          & TCSVT'23                      & VST                          & \ding{53}                      & RGB               & 0.711             & 3.030                    & 0.789             & 3.444                    & 0.728             & 5.312                    & 0.703             & 5.576                    & 0.735             & 4.341                    \\
            PAMFN \cite{zeng2024multimodal}                   & TIP'24                        & VST                          & \ding{53}                      & RGB               & 0.636             & --                       & 0.720             & --                       & 0.769             & --                       & 0.708             & --                       & 0.711             & --                       \\
            VATP-Net \cite{gedamu2024visual}                  & TCSVT'24                      & VST                          & \ding{53}                      & RGB+              & 0.800             & --                       & \textbf{0.810}    & --                       & 0.780             & --                       & 0.769             & --                       & 0.800             & --                       \\
            CoFInAl \cite{zhou2024cofinal}                    & IJCAI'24                      & VST                          & \ding{51}                      & RGB               & \underline{0.809} & \textbf{1.356}           & \underline{0.806} & 2.453                    & \underline{0.804} & 9.918                    & \textbf{0.810}    & \textbf{2.383}           & \underline{0.807} & {4.028}                  \\
            PHI (Ours)                                        & --                            & VST                          & \ding{51}                      & RGB               & \textbf{0.818}    & \underline{2.187}        & {0.803}           & \textbf{2.149}           & \textbf{0.812}    & \underline{2.119}        & \underline{0.805} & {2.744}                  & \textbf{0.810}    & \textbf{2.300}           \\
            \bottomrule
        \end{tabular}
    }
    \label{tab:sota}
\end{table*}

\section{Experiments} \label{sec_exp}
In this section, we first describe the experimental setup and then present and analyze the experimental results.

\subsection{Experimental Setups} \label{sec_setup}
\paragraph{Datasets}
In our study, we assess all models using three extensive long-term AQA datasets. The first dataset, i.e., the \textbf{Rhythmic Gymnastics (RG)} dataset \cite{zeng2020hybrid}, comprises a collection of 1,000 videos showcasing various rhythmic gymnastics actions performed with different apparatuses, including Ball, Clubs, Hoop, and Ribbon. Each video has an approximate duration of 1.6 minutes, and the frame rate is set at 25 frames per second. The dataset is split into training and evaluation sets, with 200 videos allocated for training and 50 for evaluation in each action category.
The second dataset, i.e., the \textbf{Figure Skating Video (Fis-V)} dataset \cite{pirsiavash2014assessing,parmar2017learning}, contains 500 videos capturing ladies' singles short programs in figure skating. Each video has a duration of approximately 2.9 minutes, with a frame rate set at 25 frames per second. Following the official split, the dataset is divided into 400 training videos and 100 testing videos. Each video in this dataset comes with annotations for two scores: Total Element Score (TES) and Total Program Component Score (PCS). To align with the previous method \cite{xu2019learning}, we develop separate models for different score/action types.
\linelabel{r1:C2.6-b}The third dataset, i.e., the \textbf{LOng-form GrOup (LOGO)} dataset \cite{zhang2023logo}, consists of 150 samples for training and 50 for testing.
It captures videos showcasing synchronized swimming group actions, where each video sequence is approximately 3 and a half minutes in length. Currently, LOGO has the longest video duration among all AQA datasets, making it a challenging benchmark for long-term AQA tasks.

\paragraph{Evaluation Metrics}
We use two evaluation metrics to validate the performance of all the AQA methods.

Consistent with previous long-term AQA methods \cite{xu2022likert,zeng2020hybrid}, we utilize Spearman's Rank Correlation Coefficient (SRCC) as the evaluation metric, denoted as $\rho$. The SRCC is defined as the Pearson correlation coefficient between their ranks,  $r(s_i)$ and $r(\hat{s}_i)$, to predicted and ground-truth scores, which can be formulated as follows:
\begin{equation} \label{eq_srcc}
    \rho=\frac{\sum_{i=1}^N\left(r\left(s_i\right)-\bar{r}\right)\left(r\left(\hat{s}_i\right)-\bar{r}\right)}{\sqrt{\sum_{i=1}^N\left(r\left(s_i\right)-\bar{r}\right)^2} \sqrt{\sum_{i=1}^N\left(r\left(\hat{s}_i\right)-\bar{r}\right)^2}},
\end{equation}
where $\bar{r}$ is the average rank. A higher SRCC indicates a stronger rank correlation between predicted and ground-truth scores.
Following the previous work \cite{pan2019action}, we compute the average SRCC across different action types for the RG dataset and score types for the Fis-V dataset by using Fisher’s z-value to aggregate individual SRCCs.

Compared with the previous work \cite{zhou2024cofinal}, we add a new stricter metric, the relative $\ell_2$ distance (R-$\ell_2$) \cite{yu2021group,zhou2023hierarchical}.
The purpose of introducing R-$\ell_2$ is to measure the relative error of AQA models more precisely without being affected by the score scale.
Given the highest and lowest scores for an action $s_{\mathrm{max}}$ and $s_{\mathrm{min}}$, the relative $\ell_2$ distance R-$\ell_2$ is defined as:
\begin{equation}
    \text{R-}\ell_2 = \frac{1}{N}\sum_n^N \left( \frac{|s_n - \hat{s}_n|}{s_{\mathrm{max}} - s_{\mathrm{min}}} \right)^2 \times 100,
\end{equation}
where $s_n$ and $\hat{s}_n$ represent the ground-truth score and prediction for the $n$-th sample, respectively. Fisher’s z-value is used to measure the average performance across actions.

\paragraph{Implementation Details}
We implemented PHI using PyTorch on an RTX 3090 GPU. We employ VST pre-trained on Kinetics 600 \cite{xu2022likert} and I3D on Kinetics 400 \cite{carreira2017quo} as backbones to conduct experiments, respectively. The feature dimensions $D, d_k, d_t$ are set to 1024, 128, and 32, respectively.
Following the previous work \cite{xu2022likert,zhang2023logo}, we initially partition the videos into non-overlapping 32-frame segments.
During training, we randomly determine the start segment, specifically $M=68$ for RG, $M=124$ for Fis-V, and $M=48$ for LOGO, respectively. During testing, all segments are utilized.
We optimize all models using SGD with a momentum of 0.9. The batch size $B$ is 32, and the learning rate starts at 0.01, gradually decreasing to 0.0001 through a cosine annealing strategy. For convergence, all the models are trained for 200 epochs. The loss weights $\lambda_{\mathrm{M}},\lambda_{\mathrm{R}}$ are set to 0.5 and 0.01, respectively. To further optimize the networks, we apply a dropout of 0.3 and a weight decay of 0.01.

\begin{table*}
    \centering
    \rowcolors{17}{gray!12}{gray!2}
    \setlength{\tabcolsep}{0.5em}
    \caption{
        Results of SRCC ($\uparrow$) and R-$\ell_2$ ($\downarrow$) on the Fis-V dataset.
        The best results are highlighted in \textbf{bold}, while the second-best results are \underline{underlined}.
        The symbol ``$\star$" indicates our reimplementation based on the official code.
        The average SRCC is computed using the Fisher-z value.
        ``--" denotes that the method does not report this metric.
        ``+" indicates the use of additional features/modalities.
    }
    \begin{tabular}{rrcccrrrrrr}
        \toprule
        \multirow{2}{*}[-1ex]{\bf Method}                 &
        \multirow{2}{*}[-1ex]{\bf Publisher}              &
        \multirow{2}{*}[-1ex]{\bf Backbone}               &
        \multirow{2}{*}[-1ex]{\makecell[c]{\bf Shift                                                                                                                                                                                         \\ \bf Aware}} &
        \multirow{2}{*}[-1ex]{\makecell[c]{\bf Modality}} &
        \multicolumn{2}{c}{\bf TES}                       &
        \multicolumn{2}{c}{\bf PCS}                       &
        \multicolumn{2}{c}{\bf Average}                                                                                                                                                                                                      \\
        \cmidrule(lr){6-7}\cmidrule(lr){8-9}\cmidrule(lr){10-11}
                                                          &           &      &           &      & \textbf{SRCC}     & \textbf{R-$\bm{\ell_2}$} & \textbf{SRCC}     & \textbf{R-$\bm{\ell_2}$} & \textbf{SRCC}     & \textbf{R-$\bm{\ell_2}$} \\
        \midrule
        C3D+SVR \cite{parmar2017learning}                 & TPAMI'17  & C3D  & \ding{53} & RGB  & 0.400             & --                       & 0.590             & --                       & 0.501             & --                       \\
        MS-LSTM \cite{xu2019learning}                     & TCSVT'19  & C3D  & \ding{53} & RGB  & 0.650             & --                       & 0.780             & --                       & 0.721             & --                       \\
        \hdashline
        GDLT \cite{xu2022likert}                          & CVPR'22   & I3D  & \ding{53} & RGB  & 0.260             & 5.582                    & 0.395             & 5.039                    & 0.329             & 5.311                    \\
        HGCN$^\star$ \cite{zhou2023hierarchical}          & TCSVT'23  & I3D  & \ding{53} & RGB  & 0.311             & 4.317                    & 0.407             & 4.608                    & 0.360             & 4.463                    \\
        CoFInAl \cite{zhou2024cofinal}                    & IJCAI'24  & I3D  & \ding{51} & RGB  & \underline{0.589} & \underline{3.470}        & \underline{0.788} & \textbf{2.843}           & \underline{0.702} & \underline{3.157}        \\
        PHI (Ours)                                        & --        & I3D  & \ding{51} & RGB  & \textbf{0.659}    & \textbf{2.572}           & \textbf{0.798}    & \underline{3.073}        & \textbf{0.736}    & \textbf{2.823}           \\
        \hdashline
        MS-LSTM \cite{xu2019learning}                     & TCSVT'19  & VST  & \ding{53} & RGB  & 0.660             & --                       & 0.809             & --                       & 0.744             & --                       \\
        ACTION-NET \cite{zeng2020hybrid}                  & ACM MM'20 & VST+ & \ding{53} & RGB  & 0.694             & --                       & 0.809             & --                       & 0.757             & --                       \\
        GDLT \cite{xu2022likert}                          & CVPR'22   & VST  & \ding{53} & RGB  & 0.685             & 3.717                    & 0.820             & 2.072                    & 0.761             & 2.895                    \\
        HGCN$^\star$ \cite{zhou2023hierarchical}          & TCSVT'23  & VST  & \ding{53} & RGB  & 0.246             & 12.628                   & 0.221             & 20.531                   & 0.234             & 16.580                   \\
        MLP-Mixer \cite{xia2023skating}                   & AAAI'23   & VST  & \ding{53} & RGB  & 0.680             & --                       & 0.820             & --                       & 0.750             & --                       \\
        SGN \cite{du2024learning}                         & TMM'24    & VST  & \ding{53} & RGB  & 0.700             & --                       & 0.830             & --                       & 0.765             & --                       \\
        PAMFN \cite{zeng2024multimodal}                   & TIP'24    & VST  & \ding{53} & RGB  & 0.665             & --                       & 0.823             & --                       & 0.755             & --                       \\
        VATP-Net \cite{zeng2024multimodal}                & TCSVT'24  & VST  & \ding{53} & RGB+ & 0.702             & --                       & 0.863             & --                       & 0.796             & --                       \\
        CoFInAl \cite{zhou2024cofinal}                    & IJCAI'24  & VST  & \ding{51} & RGB  & \underline{0.716} & \underline{2.875}        & \underline{0.843} & \underline{1.752}        & \underline{0.788} & \underline{2.314}        \\
        PHI (Ours)                                        & --        & VST  & \ding{51} & RGB  & \textbf{0.726}    & \textbf{2.543}           & \textbf{0.867}    & \textbf{1.656}           & \textbf{0.804}    & \textbf{2.178}           \\
        \bottomrule
    \end{tabular}
    \label{tab:sota-fis}
\end{table*}

\subsection{Results and Analysis}
We begin with a thorough comparison with state-of-the-art methods, followed by an ablation study that includes parameter sensitivity analysis, and conclude with visualizations.

\subsubsection{Comparison with the State-of-the-Art}
In our comparative analysis, we benchmarked several state-of-the-art methods, consisting of C3D+SVR \cite{parmar2017learning}, MS-LSTM \cite{xu2019learning}, ACTION-NET \cite{zeng2020hybrid}, GDLT \cite{xu2022likert}, HGCN \cite{zhou2023hierarchical}, and CoFInAl \cite{zhou2024cofinal}. The comprehensive results are reported in \cref{tab:sota,tab:sota-fis,tab:sota-logo}, and the computational performance is listed in \cref{tab:computation}.

\paragraph{Comparison with Different Backbones}
We employed backbone various architectures, namely C3D \cite{tran2015learning}, I3D \cite{carreira2017quo}, ResNet \cite{he2016deep}, and VST \cite{liu2022video}, to discern their impact on AQA performance.
Among them, I3D consistently outperformed C3D across all categories on the RG dataset, highlighting its proficiency in capturing spatial-temporal dynamics for AQA. Particularly noteworthy was the superior performance of the VST backbone, achieving the highest average SRCC on all three datasets, as can be seen in \cref{tab:sota,tab:sota-fis,tab:sota-logo}.
In the RG dataset, ACTION-NET, when coupled with the VST backbone, displayed a remarkable correlation gain of over 16.85\% compared to its I3D-based counterpart.
This highlights the VST backbone's capability to capture detailed temporal dynamics crucial for AQA.
Our proposed method, PHI, leveraging both I3D and VST backbones, showcased outstanding results across all categories, with the VST variant emerging as the top performer.
These findings validate the significance of advanced 3D convolutional architectures, such as the I3D and VST backbones, in capturing subtle action details for improved AQA performance, further enhanced by the incorporation of our PHI method.

\begin{table}
    \centering
    \rowcolors{4}{gray!2}{gray!12}
    \setlength{\tabcolsep}{0.5em}
    \caption{
        Results of SRCC ($\uparrow$) and R-$\ell_2$ ($\downarrow$) on the LOGO dataset.
        The best results are highlighted in \textbf{bold}, while the second-best results are \underline{underlined}.
    }
    \label{tab:sota-logo}
    \begin{tabular}{rrccrr}
        \toprule
        \bf Method                        & \bf Publisher & \bf Backbone & \makecell[c]{\bf Shift                                         \\ \bf Aware} & \textbf{SRCC}     & \textbf{R-$\bm{\ell_2}$}\\
        \midrule
        USDL \cite{tang2020uncertainty}   & CVPR'20       & I3D          & \ding{53}              & 0.426             & 5.736             \\
        CoRe \cite{yu2021group}           & ICCV'21       & I3D          & \ding{53}              & 0.471             & 5.402             \\
        TSA \cite{xu2022finediving}       & CVPR'22       & I3D          & \ding{53}              & 0.452             & 5.533             \\
        CoRe-GOAT \cite{zhang2023logo}    & CVPR'23       & I3D          & \ding{53}              & 0.494             & 5.072             \\
        HGCN  \cite{zhou2023hierarchical} & TCSVT'23      & I3D          & \ding{53}              & 0.471             & 4.954             \\
        CoFInAl   \cite{zhou2024cofinal}  & IJCAI'24      & I3D          & \ding{51}              & \underline{0.552} & \underline{4.586} \\
        PHI (Ours)                        & --            & I3D          & \ding{51}              & \textbf{0.713}    & \textbf{3.608}    \\
        \hdashline
        USDL \cite{tang2020uncertainty}   & CVPR'20       & VST          & \ding{53}              & 0.473             & 5.076             \\
        CoRe \cite{yu2021group}           & ICCV'21       & VST          & \ding{53}              & 0.500             & 5.960             \\
        TSA \cite{xu2022finediving}       & CVPR'22       & VST          & \ding{53}              & 0.475             & 4.778             \\
        CoRe-GOAT \cite{zhang2023logo}    & CVPR'23       & VST          & \ding{53}              & 0.560             & 4.763             \\
        HGCN  \cite{zhou2023hierarchical} & TCSVT'23      & VST          & \ding{53}              & 0.671             & 6.564             \\
        CoFInAl   \cite{zhou2024cofinal}  & IJCAI'24      & VST          & \ding{51}              & \underline{0.698} & \underline{4.019} \\
        PHI (Ours)                        & --            & VST          & \ding{51}              & \textbf{0.835}    & \textbf{2.752}    \\
        \bottomrule
    \end{tabular}
\end{table}

\paragraph{Comparison with Shift-Unaware Methods}
Traditional shift-unaware AQA methods \cite{xu2022likert,zhou2023hierarchical,xu2019learning,zeng2020hybrid} employ representation layers that may implicitly mitigate the domain shift issue. However, the persistence of domain shift often leads to suboptimal performance in \cref{tab:sota,tab:sota-fis,tab:sota-logo}.
PHI with VST consistently outperforms others across all actions and score types in all three datasets, demonstrating its effectiveness in addressing the domain shift issue in long-term AQA. Specifically, it excels in categories like Ball, Hoop, and Ribbon on RG and across TES and PCS on Fis-V by a large margin. For instance, PHI demonstrates notable performance improvements, achieving a remarkable 9.65\% and 6.14\% correlation gain in the Ball and Hoop categories, respectively, on the RG dataset compared to GDLT \cite{xu2022likert}. Overall, PHI delivers significant average correlation gains of 5.88\%, 5.65\%, and 24.44\% on RG, Fis-V, and LOGO, respectively.
These results validate the benefit of employing the two-stage instruction to address the domain shift issue.

The above statistics focus on unimodal comparisons and exclude multi-modal methods. Below, we compare PHI with recent multi-modal approaches.
\linelabel{r1:C2.5}Notably, PAMFN \cite{zeng2024multimodal} uses only RGB data, while VATP-Net \cite{gedamu2024visual} leverages multi-modal inputs. Despite being an unimodal method, PHI outperforms PAMFN (0.711) by 13.9\% and slightly surpasses the multi-modal method \cite{gedamu2023fine} on RG, as shown in \cref{tab:sota}. On Fis-V, PHI achieves an average SRCC 6.5\% higher than PAMFN (0.755) and 1.0\% higher than VATP-Net (0.796), as shown in \cref{tab:sota-fis}. These results highlight PHI’s ability to handle domain shifts without additional modalities, demonstrating that multi-modal methods do not necessarily outperform unimodal methods when domain shifts are effectively addressed.
Future work will explore extending PHI to multi-modal settings to further enhance performance and generalization.
The comparison with various state-of-the-art methods positions PHI as a promising solution for advancing AQA capabilities.

\paragraph{Comparison with Task-Level Discrepancies Solution} \label{sec:exp_task_level}
This work categorizes domain shifts into two perspectives: task-level discrepancies and feature-level discrepancies (see \cref{fig:idea-b}).
CoFInAl \cite{zhou2024cofinal} primarily addresses task-level discrepancies, whereas our work focuses on feature-level discrepancies.
Both approaches have demonstrated substantial performance improvements compared to shift-unaware methods, as validated in \cref{tab:sota,tab:sota-fis,tab:sota-logo}, demonstrating the effectiveness and necessity of explicit methods in mitigating domain shifts in long-term AQA.
\linelabel{r1:C3.5}In our experiments, we use the optimal parameters for I3D from the well-tuned parameters of VST to better validate the generalizability of the proposed method on different backbones. This means that we did not fine-tune the parameters of the I3D backbone.
As shown in \cref{tab:sota,tab:sota-fis}, PHI is still superior to CoFInAl, which has been well-tuned on the I3D backbone.
As a result, it still has significant room for improvement through further fine-tuning, which could potentially lead to even better performance. This explains why the VST backbone shows greater improvement over I3D when compared to CoFInAl.

\begin{table}
    \centering
    \rowcolors{22}{gray!2}{gray!12}
    \setlength{\tabcolsep}{0.5em}
    \caption{
        Computational comparison with existing open-source long-term AQA methods on the RG dataset.
        ``--" denotes methods without offline modules.
    }
    \begin{tabular}{rcrrrr}
        \toprule
        \multirow{2}{*}[-1ex]{\bf Method}     &
        \multirow{2}{*}[-1ex]{\makecell{\bf Shift                                                                      \\ \bf Aware}} &
        \multirow{2}{*}[-1ex]{\makecell{\bf FLOPs                                                                      \\\bf(G)}} &
        \multicolumn{2}{c}{\bf Parameter (M)} &
        \multirow{2}{*}[-1ex]{\makecell{\bf Inference                                                                  \\\bf Time (ms)}} \\
        \cmidrule(lr){4-5}
                                              &           &          & \textbf{Online} & \textbf{Offline} &            \\
        \midrule
        ACTION-NET \cite{zeng2020hybrid}      & \ding{53} & 34.7500  & 28.08           & --               & 305.2474   \\
        GDLT \cite{xu2022likert}              & \ding{53} & ~~0.1164 & ~~3.20          & --               & ~~~~3.2249 \\
        HGCN \cite{zhou2023hierarchical}      & \ding{53} & ~~1.1201 & ~~0.50          & --               & ~~~~6.7830 \\
        CoFInAl \cite{zhou2024cofinal}        & \ding{51} & ~~0.1178 & ~~3.70          & --               & ~~~~3.8834 \\
        PHI-half (Ours)                       & \ding{51} & ~~0.2637 & ~~2.80          & ~~4.60           & ~~~~5.6870 \\
        PHI (Ours)                            & \ding{51} & ~~0.2637 & ~~3.00          & ~~4.60           & ~~~~5.6870 \\
        \bottomrule
    \end{tabular}
    \label{tab:computation}
\end{table}

\begin{table*}
    \centering
    \caption{Ablation results on the RG dataset.}
    \rowcolors{8}{gray!2}{gray!12}
    \setlength{\tabcolsep}{0.5em}
    \resizebox{\linewidth}{!}{
        \begin{tabular}{llllllllllll}
            \toprule
            \multirow{2}{*}[-0.5ex]{\textbf{Setting}} & \multicolumn{2}{c}{\textbf{Ball}} & \multicolumn{2}{c}{\textbf{Clubs}}           & \multicolumn{2}{c}{\textbf{Hoop}} & \multicolumn{2}{c}{\textbf{Ribbon}} & \multicolumn{2}{c}{\textbf{Average}}                                         \\
            \cmidrule(lr){2-3} \cmidrule(lr){4-5} \cmidrule(lr){6-7} \cmidrule(lr){8-9} \cmidrule(lr){10-11}
                                                      & \multicolumn{1}{c}{\textbf{SRCC}} & \multicolumn{1}{c}{\textbf{R-$\bm{\ell_2}$}}
                                                      & \multicolumn{1}{c}{\textbf{SRCC}} & \multicolumn{1}{c}{\textbf{R-$\bm{\ell_2}$}}
                                                      & \multicolumn{1}{c}{\textbf{SRCC}} & \multicolumn{1}{c}{\textbf{R-$\bm{\ell_2}$}}
                                                      & \multicolumn{1}{c}{\textbf{SRCC}} & \multicolumn{1}{c}{\textbf{R-$\bm{\ell_2}$}}
                                                      & \multicolumn{1}{c}{\textbf{SRCC}} & \multicolumn{1}{c}{\textbf{R-$\bm{\ell_2}$}}                                                                                                                                                          \\
            \midrule
            PHI                                       & 0.818                             & 2.187                                        & 0.803                             & 2.149                               & 0.812                                & 2.199 & 0.805 & 2.744 & 0.810 & 2.300 \\
            \hdashline
            PHI-half
                                                      & 0.808 $^{\downarrow 1.22\%}$      & 2.027 $^{\downarrow 7.32\%}$
                                                      & 0.790 $^{\downarrow 1.62\%}$      & 2.460 $^{\uparrow 14.47\%}$
                                                      & 0.788 $^{\downarrow 2.96\%}$      & 5.776 $^{\uparrow 162.80\%}$
                                                      & 0.752 $^{\downarrow 6.58\%}$      & 2.900 $^{\uparrow 5.68\%}$
                                                      & 0.785 $^{\downarrow 3.09\%}$      & 3.291 $^{\uparrow 43.00\%}$                                                                                                                                                                           \\
            ~~w/o GMF
                                                      & 0.802 $^{\downarrow 1.96\%}$      & 1.942 $^{\downarrow 11.20\%}$
                                                      & 0.784 $^{\downarrow 2.36\%}$      & 2.316 $^{\uparrow 7.77\%}$
                                                      & 0.789 $^{\downarrow 2.83\%}$      & 6.733 $^{\uparrow 206.27\%}$
                                                      & 0.756 $^{\downarrow 6.09\%}$      & 3.217 $^{\uparrow 17.24\%}$
                                                      & 0.783 $^{\downarrow 3.33\%}$      & 3.552 $^{\uparrow 54.43\%}$                                                                                                                                                                           \\
            ~~w/o TESA
                                                      & 0.661 $^{\downarrow 19.20\%}$     & 3.347 $^{\uparrow 53.09\%}$
                                                      & 0.637 $^{\downarrow 20.66\%}$     & 3.904 $^{\uparrow 81.63\%}$
                                                      & 0.371 $^{\downarrow 54.32\%}$     & 6.490 $^{\uparrow 195.13\%}$
                                                      & 0.550 $^{\downarrow 31.68\%}$     & 5.016 $^{\uparrow 82.91\%}$
                                                      & 0.564 $^{\downarrow 30.37\%}$     & 4.689 $^{\uparrow 103.87\%}$                                                                                                                                                                          \\
            ~~w/o LCR
                                                      & 0.775 $^{\downarrow 5.26\%}$      & 3.105 $^{\uparrow 42.00\%}$
                                                      & 0.773 $^{\downarrow 3.74\%}$      & 2.588 $^{\uparrow 20.40\%}$
                                                      & 0.789 $^{\downarrow 2.83\%}$      & 7.367 $^{\uparrow 235.09\%}$
                                                      & 0.719 $^{\downarrow 10.68\%}$     & 3.671 $^{\uparrow 33.81\%}$
                                                      & 0.765 $^{\downarrow 5.56\%}$      & 4.183 $^{\uparrow 81.87\%}$                                                                                                                                                                           \\
            ~~w/o KL
                                                      & 0.806 $^{\downarrow 1.47\%}$      & 1.991 $^{\downarrow 8.96\%}$
                                                      & 0.799 $^{\downarrow 0.50\%}$      & 2.231 $^{\uparrow 3.81\%}$
                                                      & 0.777 $^{\downarrow 4.31\%}$      & 7.005 $^{\uparrow 218.56\%}$
                                                      & 0.778 $^{\downarrow 3.35\%}$      & 3.174 $^{\uparrow 15.68\%}$
                                                      & 0.790 $^{\downarrow 2.47\%}$      & 3.600 $^{\uparrow 56.52\%}$                                                                                                                                                                           \\
            \bottomrule
        \end{tabular}
    }
    \label{tab:ablation}
\end{table*}

\begin{table*}
    \centering
    \caption{Ablation results on the Fis-V dataset.}
    \rowcolors{6}{gray!2}{gray!12}
    \setlength{\tabcolsep}{0.5em}
    \resizebox{0.75\linewidth}{!}{
        \begin{tabular}{llllllll}
            \toprule
            \multirow{2}{*}[-0.5ex]{\textbf{Setting}} & \multicolumn{2}{c}{\textbf{TES}}  & \multicolumn{2}{c}{\textbf{PCS}}             & \multicolumn{2}{c}{\textbf{Average}}                                                                                               \\
            \cmidrule(lr){2-3} \cmidrule(lr){4-5} \cmidrule(lr){6-7}
                                                      & \multicolumn{1}{c}{\textbf{SRCC}} & \multicolumn{1}{c}{\textbf{R-$\bm{\ell_2}$}}
                                                      & \multicolumn{1}{c}{\textbf{SRCC}} & \multicolumn{1}{c}{\textbf{R-$\bm{\ell_2}$}}
                                                      & \multicolumn{1}{c}{\textbf{SRCC}} & \multicolumn{1}{c}{\textbf{R-$\bm{\ell_2}$}}                                                                                                                                      \\
            \midrule
            PHI                                       & 0.726                             & 2.543                                        & 0.867                                & 1.656                        & 0.804                         & 2.178                        \\
            \hdashline
            PHI-half                                  & 0.661 $^{\downarrow 9.07\%}$      & 3.050 $^{\uparrow 19.88\%}$                  & 0.861 $^{\downarrow 1.15\%}$         & 1.183 $^{\uparrow 28.58\%}$  & 0.780 $^{\downarrow 3.48\%}$  & 2.117 $^{\uparrow 2.79\%}$   \\
            ~~w/o GMF                                 & 0.668 $^{\downarrow 8.06\%}$      & 3.413 $^{\uparrow 34.33\%}$                  & 0.857 $^{\downarrow 0.58\%}$         & 1.769 $^{\uparrow 6.72\%}$   & 0.780 $^{\downarrow 3.48\%}$  & 2.591 $^{\uparrow 18.97\%}$  \\
            ~~w/o TESA                                & 0.627 $^{\downarrow 13.87\%}$     & 4.360 $^{\uparrow 166.33\%}$                 & 0.776 $^{\downarrow 10.97\%}$        & 3.714 $^{\uparrow 123.67\%}$ & 0.709 $^{\downarrow 11.84\%}$ & 4.037 $^{\uparrow 84.81\%}$  \\
            ~~w/o LCR                                 & 0.280 $^{\downarrow 61.40\%}$     & 5.576 $^{\uparrow 238.17\%}$                 & 0.294 $^{\downarrow 66.16\%}$        & 4.984 $^{\uparrow 200.91\%}$ & 0.282 $^{\downarrow 65.00\%}$ & 5.280 $^{\uparrow 142.05\%}$ \\
            ~~w/o KL                                  & 0.576 $^{\downarrow 20.93\%}$     & 2.437 $^{\downarrow 3.88\%}$                 & 0.841 $^{\downarrow 2.56\%}$         & 2.564 $^{\uparrow 54.94\%}$  & 0.780 $^{\downarrow 3.48\%}$  & 2.501 $^{\uparrow 14.74\%}$  \\
            \bottomrule
        \end{tabular}
    }
    \label{tab:ablation-supp}
\end{table*}

Although PHI is weaker than CoFInAl in some categories, PHI exhibits a 2.03\% performance gain on the challenging Fis-V dataset, while achieving a modest 0.37\% gain on the simpler RG dataset.
\linelabel{r1:C2.4}\linelabel{r1:C3.5-b}The smaller improvement observed on the RG dataset can be attributed to two key factors: its limited sample size and the relatively mild domain shift between the pre-training task and AQA. Specifically, RG comprises a smaller dataset compared to Fis-V (refer to the dataset details in \cref{sec_setup}), which restricts the model's capacity to learn complex domain adaptation patterns.
Furthermore, as illustrated in \cref{fig:tsne_ball-a}, RG exhibits less pronounced feature misalignment, reducing the upper limit of performance improvement for extensive domain adaptation.
In contrast, as can be seen in \cref{fig:tsne-a}, Fis-V demonstrates significant domain discrepancies, where PHI's domain shift mitigation mechanisms prove more impactful, resulting in a more substantial performance gain.
Notably, PHI achieves a significant 42.9\% reduction in R-$\ell_2$ error on RG, indicating PHI's robustness in effectively managing subtle feature discrepancies.
Furthermore, the performance gap in CoFInAl can be attributed to the discretization of the continuous score space for classification.
Our novel contribution focuses on addressing feature-level discrepancies in AQA, which we have identified as crucial for achieving additional performance improvements, compared to the task-level one. By innovatively tackling these discrepancies, the robustness and accuracy of AQA have been enhanced.

We further compare and analyze both solutions.
While CoFInAl and PHI address the domain shift in complementary ways, they are fundamentally incompatible due to their opposing alignment directions.
\linelabel{r1:C3.6}As can be seen from the different arrows in \cref{fig:idea-b}, these two approaches operate the alignment process in opposite directions, making their direct combination infeasible.
If pre-trained features are already well-optimized and generalizable for AQA, task-level alignment (CoFInAl) may be sufficient.
However, our experimental results in \cref{tab:sota,tab:sota-fis,tab:sota-logo} demonstrate that additional performance gains are achievable through PHI, indicating that pre-trained features still require adaptation to better suit AQA.
In the future, we aim to explore the potential of combining both approaches to fully leverage their respective strengths and achieve even higher performance.
Moreover, PHI is designed as a modular enhancement and can be integrated into other baseline methods that do not explicitly consider domain shift. By refining feature alignment, PHI enhances model robustness and generalization, making it a versatile tool for improving AQA performance across various architectures.

\paragraph{Computational Efficiency and Model Complexity}
To provide a more comprehensive comparison, we evaluated PHI and existing long-term AQA methods under identical conditions, with a focus on computational efficiency. The results are summarized in \cref{tab:computation}.
PHI consists of online and offline components, strategically designed to balance computational efficiency with assessment accuracy. The online module (flow path) handles real-time inference with minimal latency, while the offline module (TETE) refines feature representations, thus reducing the computational burden during online inference. Combining the results from \cref{tab:sota,tab:sota-fis,tab:sota-logo}, the distillation design enables it to achieve a competitive balance between performance and efficiency compared to the state-of-the-art.

\linelabel{r1:C1.4}Notably, CoFInAl achieves the lowest FLOPs (0.1178G) and inference time (3.8834ms) among shift-aware methods, underscoring its computational efficiency. However, PHI compensates for a slightly higher computational cost by leveraging a progressive instruction strategy. This approach improves assessment accuracy with only a 0.1470G increase in FLOPs and an additional 2.4621ms inference delay compared to GDLT.
Despite this, PHI outperforms in terms of online parameters and quality assessment performance, while maintaining a reasonable computational footprint.
PHI introduces only a slight increase in FLOPs and parameters, while significantly improving the assessment accuracy. PHI reduces FLOPs (by 99.2\%), parameters (by 89.3\%), and inference time (by 98.1\%) compared to ACTION-NET, and also achieves lower FLOPs and competitive inference time relative to HGCN. Additionally, PHI benefits from an offline distillation mechanism, allowing the TETE module to transfer knowledge to a lightweight flow model, further reducing online parameters and computational cost.
In summary, while CoFInAl excels in computational efficiency, PHI strikes an optimal balance between computational cost and assessment accuracy, making it a practical solution for practical AQA applications.

\subsubsection{Ablation Studies}
\begin{figure}
    \centering
    \includegraphics[width=\linewidth]{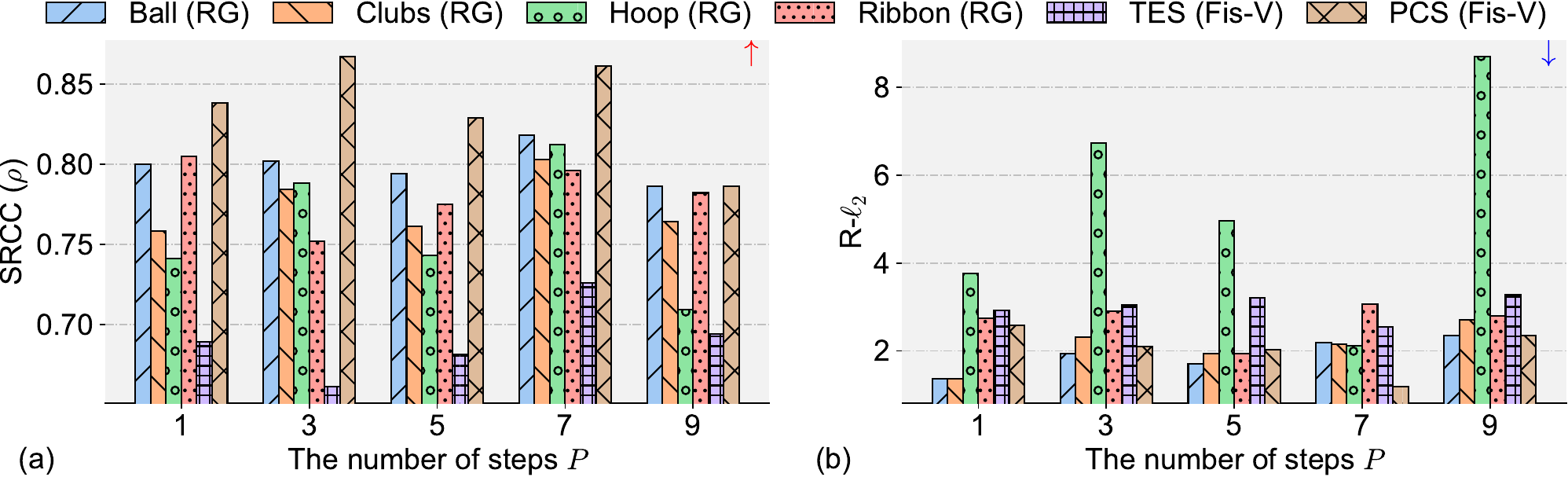}
    \caption{Results of (a) SRCC and (b) R-$\ell_2$ on the impact of different steps. The symbol ``$\uparrow$" indicates higher is better, while the symbol ``$\downarrow$" indicates lower is better.}
    \label{fig:step}
\end{figure}

This study aims to investigate the individual contributions of the core components of PHI, particularly focusing on GMF (see \cref{h1}), TESA (see \cref{sec_s2d-att}), LCR (see \cref{h2}), and the use of KL divergence loss (see \cref{eqcr}). \cref{tab:ablation,tab:ablation-supp,fig:step} present the results on RG and Fis-V.

\paragraph{Validation of Hypothesis 1}
To validate the effectiveness of \cref{h1}, we compare removing GMF (\textit{w/o GMF}) and TETE (\textit{w/o TETE}), respectively.
On the one hand, when removing GMF while retaining only TETE (\textit{w/o GMF}), we observe a decrease in performance across all categories, with an average SRCC decrease of 3.33\% and an average R-$\ell_2$ increase of 54.43\% on the RG dataset (see \cref{tab:ablation}) and an average SRCC decrease of 3.48\% and an average R-$\ell_2$ increase of 18.97\% on the Fis-V dataset (see \cref{tab:ablation-supp}).
\linelabel{r1:C2.2}These results indicate that the desired features estimated by TETE alone contain inaccuracies due to the inherent domain gap. The GMF module plays a crucial role in refining these estimations by progressively reducing the domain shift, thereby enhancing the reliability of feature representations and improving action assessment performance.
On the one hand, by replacing TESA with vanilla attention (\textit{w/o TESA}), we observe a more substantial decrease in performance, with an average SRCC decrease of 30.37\% on the RG dataset (see \cref{tab:ablation}) and an average SRCC decrease of 11.84\% on the Fis-V dataset (see \cref{tab:ablation-supp}). This emphasizes the critical role of TESA in capturing long-range dependencies essential for accurate AQA.
These results collectively validate the effectiveness of \cref{h1}

\paragraph{Validation of Hypothesis 2}
To validate the effectiveness of \cref{h2}, we compare removing LCR (\textit{w/o LCR}) and KL (\textit{w/o KL}), respectively.
On the one hand, removing LCR (\textit{w/o LCR}) results in a notable decrease in performance, with an average SRCC decrease of 5.56\% and an average R-$\ell_2$ increase of 81.87\% on the RG dataset (see \cref{tab:ablation}) and an average SRCC decrease of 65.00\% and an average R-$\ell_2$ increase of 142.15\% on the Fis-V dataset (see \cref{tab:ablation-supp}).
This demonstrates that LCR plays a crucial role in learning representations focusing on fine-grained cues, which are vital for mitigating domain shift and improving AQA performance.
On the one hand, replacing the KL divergence loss with MSE (\textit{w/o KL}) leads to a significant decrease in performance, with an average R-$\ell_2$ increase of 56.52\% on the RG dataset (see \cref{tab:ablation}) and an average R-$\ell_2$ increase of 14.74\% on the Fis-V dataset (see \cref{tab:ablation-supp}). This highlights the effectiveness of the KL divergence loss in guiding the model to learn more robust representations aligned with AQA.
These results collectively validate the effectiveness of \cref{h2}.

\paragraph{Impact of the Model Size}
As shown in \cref{tab:ablation,tab:ablation-supp}, reducing the parameter size of the flow network $\phi$ by half (PHI-half) results in a noticeable decrease in performance across all categories. Specifically, we observe an average decrease of 3.09\% in SRCC and an average increase of 43.00\% in R-$\ell_2$ on the RG dataset and an average decrease of 3.48\% in SRCC and an average increase of 2.79\% in R-$\ell_2$ on the RG dataset. Combined with the reported results in \cref{tab:sota,tab:sota-fis}, we observe that PHI-half still outperforms some strong baselines \cite{xu2022likert,zhou2023hierarchical}. This finding shows the importance of model size in maintaining performance levels, suggesting that a larger parameter space with simple MLPs contributes to better overall performance, indicating the effectiveness of PHI.

\begin{figure}
    \centering
    \includegraphics[width=\linewidth,clip,trim=280 215 110 230]{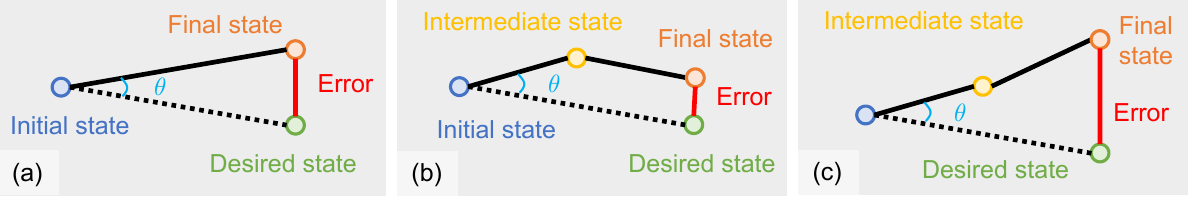}
    \caption{
        Conceptual illustrations of (a) one-step and (b,c) multi-step flows. In (a), the one-step flow allows for a fast and direct reduction of domain gaps, making it computationally efficient. However, it may lead to larger errors. Suppose the second state deviates from the desired direction with the same degree $\theta$.  In contrast, the multi-step flow (b) provides a more robust alignment through gradual refinements, reducing the risk of large errors. However, in certain cases (c), multi-step alignment can accumulate errors, potentially leading to worse performance than the one-step approach.
    }
    \label{fig:one_step}
    {
        \phantomsubcaption\label{fig:one_step-a}
        \phantomsubcaption\label{fig:one_step-b}
        \phantomsubcaption\label{fig:one_step-c}
    }
\end{figure}

\paragraph{Impact of Different Steps}
The number of flow steps plays a crucial role in refining pre-trained features for AQA. Adjusting this parameter influences the model’s ability to reduce the domain gap between pre-trained features and AQA tasks (see \cref{fig:step}).
\linelabel{r1:C3.4}Firstly, while increasing the number of steps yields slight improvements, the gains are not always substantial.
\linelabel{r1:C3.4-b}To provide an intuitive understanding, \cref{fig:one_step} illustrates the conceptual differences between one-step and multi-step flows. The one-step approach (see \cref{fig:one_step-a}) achieves competitive performance due to its direct alignment but is more susceptible to deviations. In contrast, the multi-step flow (see \cref{fig:one_step-b}) provides more controlled and gradual refinement, potentially reducing large alignment errors.
However, multi-step alignment can also introduce cumulative errors (see \cref{fig:one_step-c}), especially when the initial step already aligns features effectively. This explains why, in some cases, multi-step flows offer only marginal improvements or even perform worse than the one-step approach (in \cref{fig:step}). Nonetheless, multi-step alignment enhances stability and robustness, particularly in scenarios where direct alignment might lead to suboptimal feature adaptation.
Additionally, increasing the number of steps allows for a more gradual transformation of initial features into task-specific representations, potentially improving accuracy and reliability. However, this comes at the cost of higher computational complexity and longer training time. Conversely, reducing the number of steps accelerates training but may limit the model’s ability to effectively align with AQA tasks.
Importantly, our method is designed to support both one-step and multi-step alignment, offering flexibility depending on the computational constraints and accuracy requirements. The absence of a clear performance trend across different step settings further highlights the robustness of PHI.

\begin{table}[!t]
    \centering
    \caption{Results of different training strategies.}
    \rowcolors{2}{gray!2}{gray!12}
    \label{tab_training}
    \begin{tabular}{cll}
        \toprule
        \bf Strategy & \bf RG                       & \bf Fis-V                    \\ \midrule
        Two-stage    & 0.810                        & 0.804                        \\
        One-stage    & 0.807 $^{\downarrow 0.37\%}$ & 0.800 $^{\downarrow 0.50\%}$ \\
        \bottomrule
    \end{tabular}
\end{table}

\paragraph{Impact of Different Training Strategies}
We adopt a two-stage training process in our approach. Initially, we train the TETE to refine the estimation of desired features, focusing on obtaining accurate representations essential for the flow model. Subsequently, we integrate these refined features into our overall framework, enabling joint training with components like GMF.
To demonstrate the efficacy of the two-stage training approach, we conducted experiments comparing it to a one-stage joint training process. The results, shown in \cref{tab_training}, reveal that the two-stage process outperforms the one-stage process, with improvements of 0.37\% on the RG dataset and 0.50\% on the Fis-V dataset. This highlights the clear advantages of the two-stage training strategy.

\subsubsection{Qualitative and Quantitative Results}
We present a diverse range of visualizations to demonstrate both the qualitative and quantitative performance of our method.

\begin{figure}
    \centering
    \includegraphics[width=\linewidth]{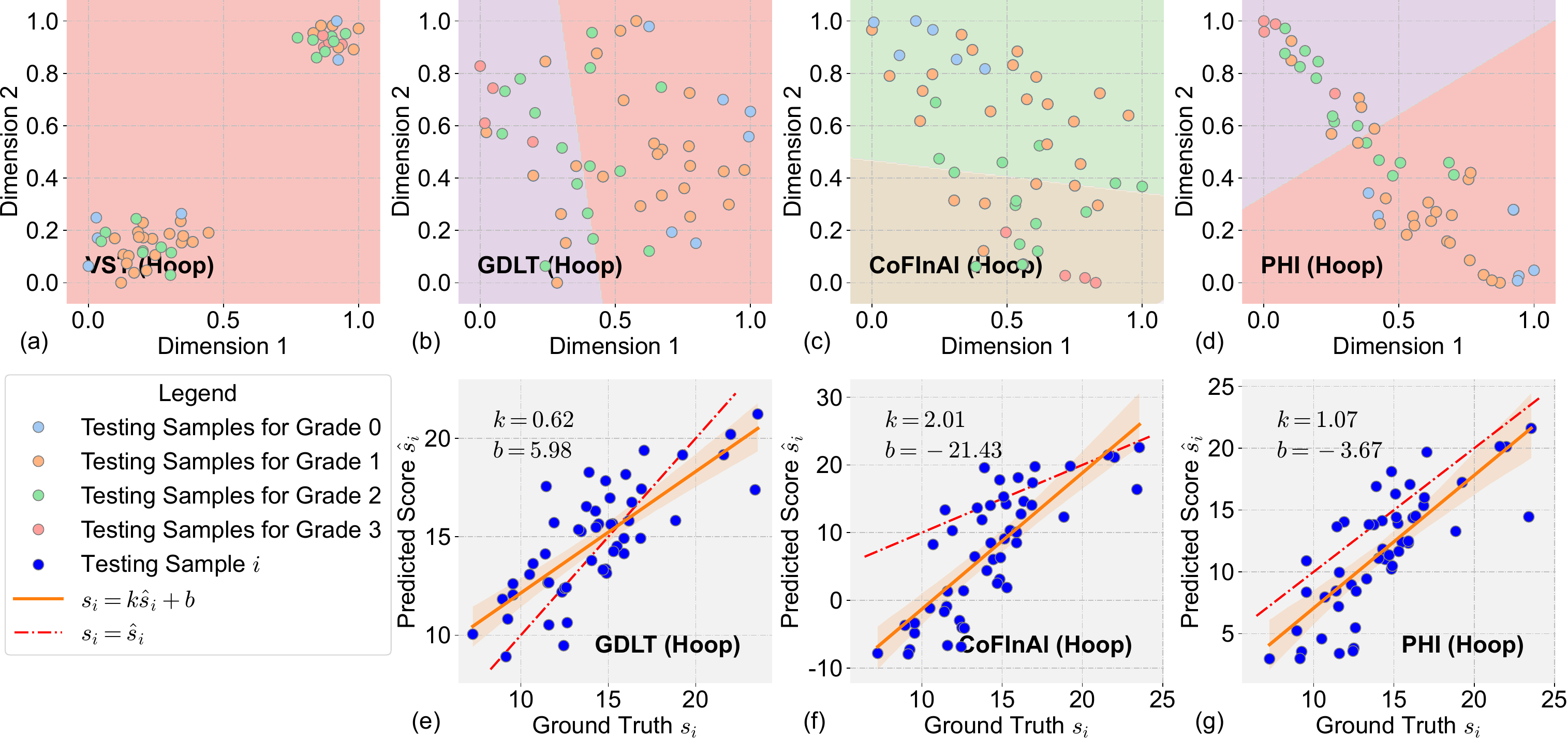}
    \caption{
        Visualization depicting the mitigation of domain shift on the Hoop (RG) dataset: t-SNE feature distribution plots (a, b, c, d) and correlation comparison plots (e, f, g) of GDLT \cite{xu2022likert}, CoFInAl \cite{zhou2024cofinal}, and our PHI method.
        The dataset is split into four grades.
    }
    \label{fig:tsne_ball}
    {
        \phantomsubcaption\label{fig:tsne_ball-a}
        \phantomsubcaption\label{fig:tsne_ball-b}
        \phantomsubcaption\label{fig:tsne_ball-c}
        \phantomsubcaption\label{fig:tsne_ball-d}
        \phantomsubcaption\label{fig:tsne_ball-e}
        \phantomsubcaption\label{fig:tsne_ball-f}
        \phantomsubcaption\label{fig:tsne_ball-h}
        \phantomsubcaption\label{fig:tsne_ball-g}
    }
\end{figure}

\begin{figure}
    \centering
    \begin{overpic}[clip,width=\linewidth]{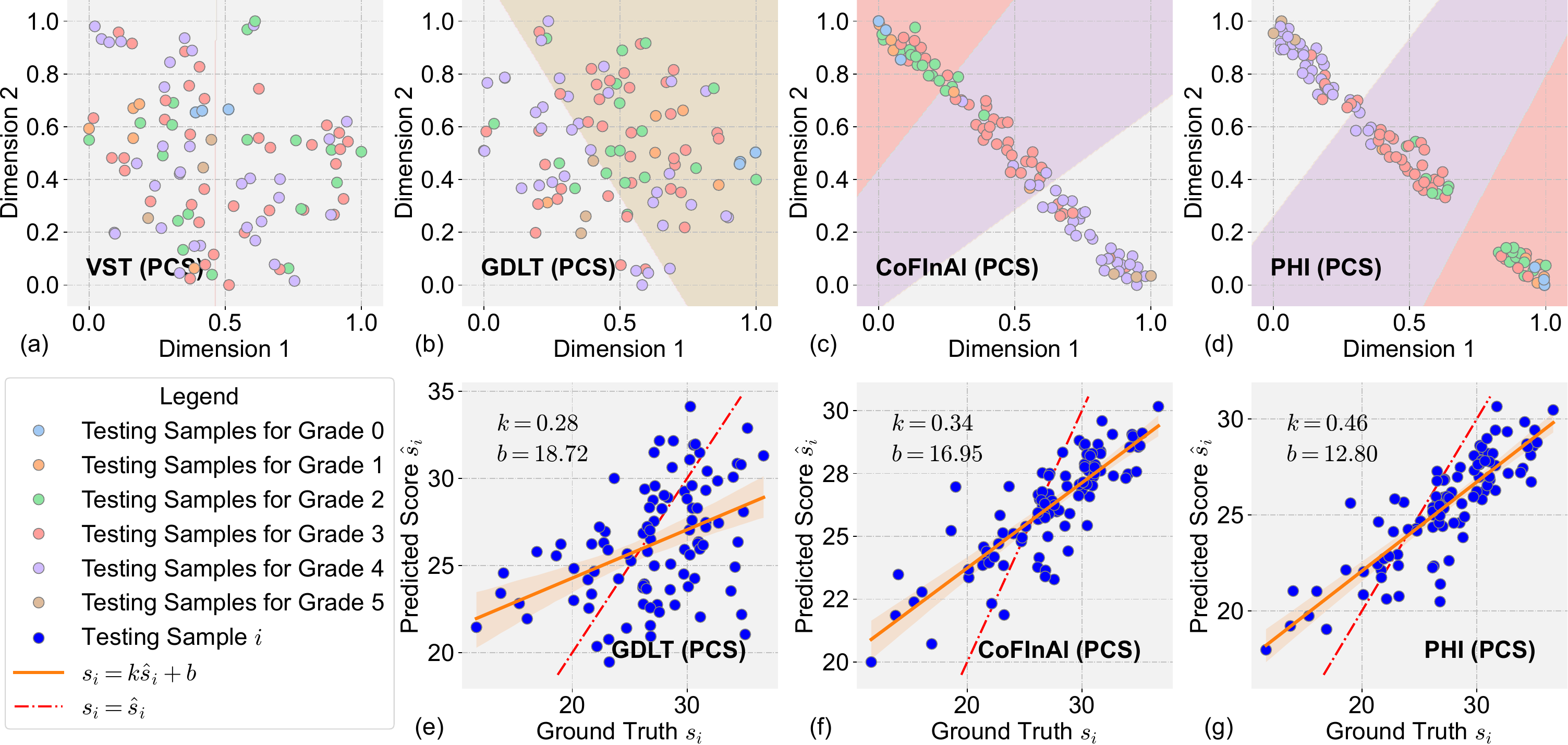}
    \end{overpic}
    \caption{
        Visualization depicting the mitigation of domain shift on the PCS (Fis-V) dataset: t-SNE feature distribution plots (a, b, c, d) and correlation comparison plots (e, f, g) of GDLT \cite{xu2022likert}, CoFInAl \cite{zhou2024cofinal}, and our PHI method.
        The dataset is split into six grades.
    }
    \label{fig:tsne}
    {
        \phantomsubcaption\label{fig:tsne-a}
        \phantomsubcaption\label{fig:tsne-b}
        \phantomsubcaption\label{fig:tsne-c}
        \phantomsubcaption\label{fig:tsne-d}
        \phantomsubcaption\label{fig:tsne-e}
        \phantomsubcaption\label{fig:tsne-f}
        \phantomsubcaption\label{fig:tsne-h}
        \phantomsubcaption\label{fig:tsne-g}
    }
\end{figure}

\paragraph{Visualization of the Domain Shift}
We visually demonstrate the effectiveness of our PHI method in mitigating domain shifts through feature distribution visualizations in the latent space and correlation analyses between predicted and ground truth scores. Specifically, we employ the t-SNE toolbox to generate feature distribution plots, which are presented in the first rows of \cref{fig:tsne_ball,fig:tsne}. Additionally, we illustrate the correlation between predicted and actual scores through scatter plots, as shown in the second rows of \cref{fig:tsne_ball,fig:tsne}.
\linelabel{r1:C3.3}To assess the model’s generalization capability, we visualize two action categories, Hoop and PCS, selected from the RG and PCS datasets, respectively. In the feature distribution plots, different score ranges (or grades) are delineated using an SVC classifier. Due to variations in dataset scales, we categorize the test samples into four score ranges for Hoop (RG), assigning labels from 0 to 3, and six score ranges for PCS (Fis-V), assigning labels from 0 to 5. Improved feature clustering, where samples of the same grade (represented by similar color shading) occupy the same region, indicates a more structured feature space for action assessment. Furthermore, we provide comparisons with GDLT \cite{xu2022likert} and CoFInAl \cite{zhou2024cofinal} to highlight the advantages of our approach.
Since the Fis-V dataset contains a larger number of test samples compared to RG, it provides a more robust and reliable evaluation of model performance and generalizability. Therefore, we primarily focus on comparisons conducted on the  PCS (Fis-V) dataset for a more comprehensive assessment as below.

Specifically, the features extracted by the VST backbone, as shown in \cref{fig:tsne-a}, exhibit a mixed distribution, indicating difficulties in distinguishing between different score ranges. This suggests that the pre-trained backbone may not be well-suited for the AQA task, leading to significant domain shift issues. In \cref{fig:tsne-b}, the feature distribution of GDLT appears confused, reflecting ineffective feature learning. In contrast, \cref{fig:tsne-c,fig:tsne-d} illustrate the feature distributions of CoFInAl and PHI, respectively, which display more distinct clustering. This clearer separation facilitates the identification of samples within each score range, indicating that both CoFInAl and PHI effectively mitigate domain shift.
While the advantage of PHI can be observed by comparing the feature plots in \cref{fig:tsne_ball-c,fig:tsne_ball-d}, the t-SNE visualizations in \cref{fig:tsne-c,fig:tsne-d} do not provide a definitive comparison between the domain shift mitigation capabilities of PHI and CoFInAl. Notably, CoFInAl suffers from a loss of precision in score prediction, which affects its fine-grained assessment capability. In contrast, PHI maintains higher precision, leading to improved reliability in action assessment. Below, we further verify that PHI outperforms CoFInAl.

\cref{fig:tsne-e,fig:tsne-f,fig:tsne-g} compare correlation plots between GDLT, CoFInAl, and PHI. In \cref{fig:tsne-e}, the correlation line ($\hat{s}_i=0.28 s_i + 18.72$) of GDLT shows a deviation from the ideal correlation line ($\hat{s}_i=s_i$), indicating a weak correlation with ground truth scores.
In \cref{fig:tsne-f}, the correlation line ($\hat{s}_i=0.34 s_i + 16.95$) of CoFInAl shows a smaller deviation from the ideal line, suggesting a stronger correlation with ground truth scores compared to GDLT. Notably, \cref{fig:tsne-g} shows that the correlation line of PHI ($\hat{s}_i=0.46 s_i + 12.80$) is the closest to the ideal line, demonstrating the highest correlation with ground truth scores among the three methods.
This suggests that PHI achieves a higher correlation with ground truth scores compared to GDLT and CoFInAl, further highlighting its superiority in mitigating domain shift and improving AQA performance.
Overall, the visualizations in \cref{fig:tsne} provide compelling evidence of PHI's effectiveness in mitigating domain shift and improving long-range AQA performance compared to existing methods.

\begin{figure*}
    \centering
    \setlength{\tabcolsep}{0em}
    \begin{minipage}{0.49\linewidth}
        \begin{tabular}{m{0.03\linewidth}<{\centering}m{0.97\linewidth}<{\centering}}
            \multicolumn{2}{c}{
                \begin{overpic}[clip,width=0.98\linewidth]{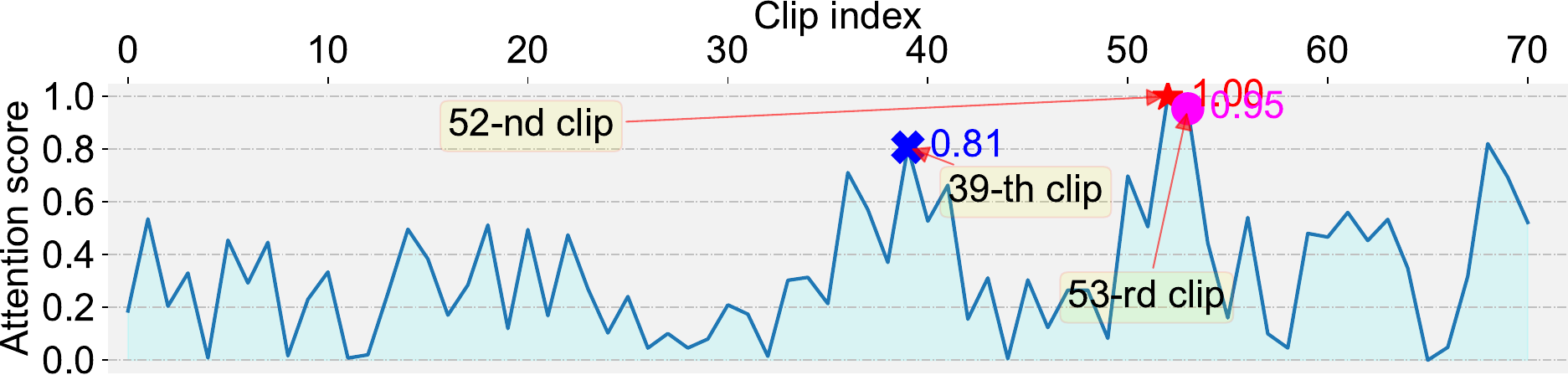}
                \end{overpic}
            }
            \\
            \rotatebox{90}{\scriptsize \color{blue} 39${\text{-th}}$ clip}
             &
            \begin{overpic}[clip,width=0.1850\linewidth]{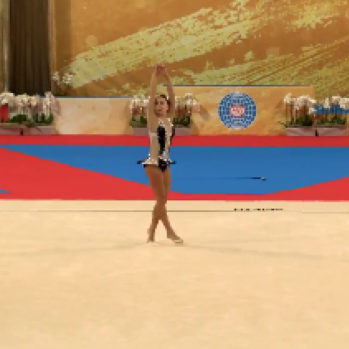}
                \put(3,5){\tiny\color{blue}1248${\text{-th}}$ frame}
            \end{overpic}
            \begin{overpic}[clip,width=0.1850\linewidth]{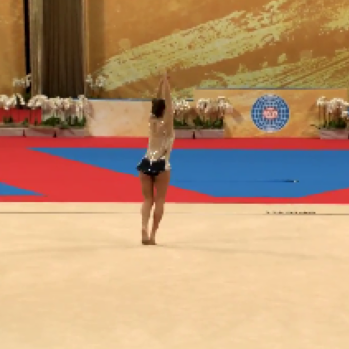}
                \put(3,5){\tiny\color{blue}1254${\text{-th}}$ frame}
            \end{overpic}
            \begin{overpic}[clip,width=0.1850\linewidth]{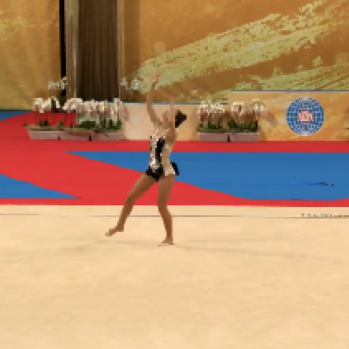}
                \put(3,5){\tiny\color{blue}1260${\text{-th}}$ frame}
            \end{overpic}
            \begin{overpic}[clip,width=0.1850\linewidth]{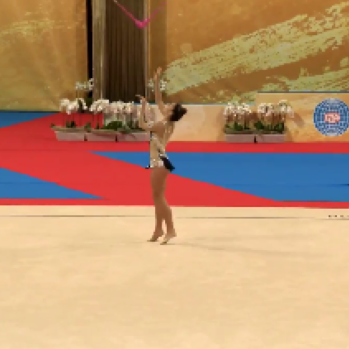}
                \put(3,5){\tiny\color{blue}1266${\text{-th}}$ frame}
            \end{overpic}
            \begin{overpic}[clip,width=0.1850\linewidth]{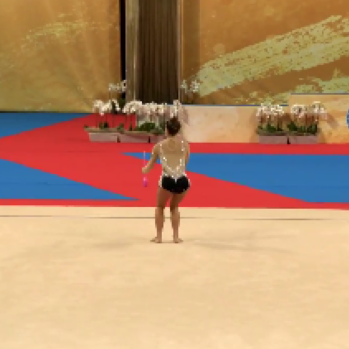}
                \put(3,5){\tiny\color{blue}1271${\text{-st}}$ frame}
            \end{overpic}
            \\
            \rotatebox{90}{\scriptsize \color{red} 52${\text{-nd}}$ clip}
             &
            \begin{overpic}[clip,width=0.1850\linewidth]{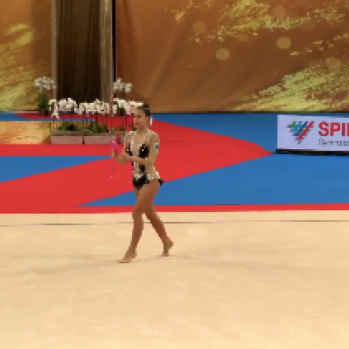}
                \put(3,5){\tiny\color{red}1664${\text{-th}}$ frame}
            \end{overpic}
            \begin{overpic}[clip,width=0.1850\linewidth]{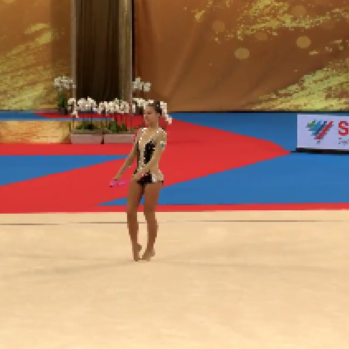}
                \put(3,5){\tiny\color{red}1670${\text{-th}}$ frame}
            \end{overpic}
            \begin{overpic}[clip,width=0.1850\linewidth]{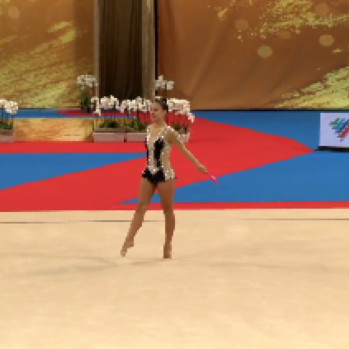}
                \put(3,5){\tiny\color{red}1676${\text{-th}}$ frame}
            \end{overpic}
            \begin{overpic}[clip,width=0.1850\linewidth]{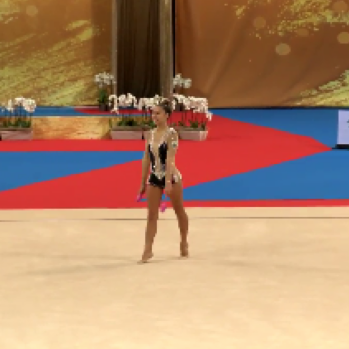}
                \put(3,5){\tiny\color{red}1681${\text{-st}}$ frame}
            \end{overpic}
            \begin{overpic}[clip,width=0.1850\linewidth]{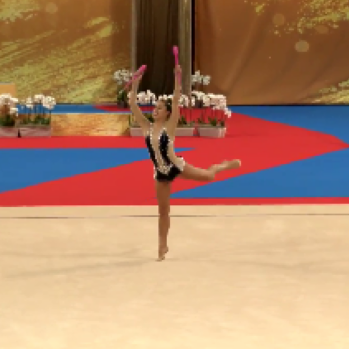}
                \put(3,5){\tiny\color{red}1688${\text{-th}}$ frame}
            \end{overpic}
            \\
            \rotatebox{90}{\scriptsize \color{magenta} 53${\text{-rd}}$ clip}
             &
            \begin{overpic}[clip,width=0.1850\linewidth]{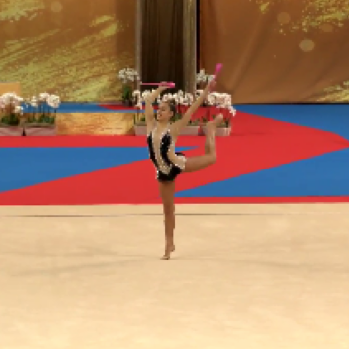}
                \put(3,5){\tiny\color{magenta}1696${\text{-th}}$ frame}
            \end{overpic}
            \begin{overpic}[clip,width=0.1850\linewidth]{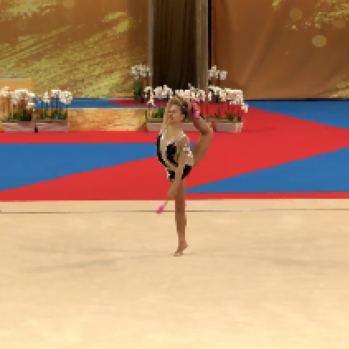}
                \put(3,5){\tiny\color{magenta}1701${\text{-st}}$ frame}
            \end{overpic}
            \begin{overpic}[clip,width=0.1850\linewidth]{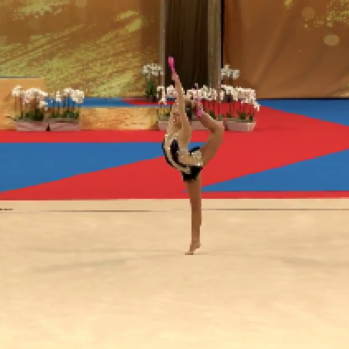}
                \put(3,5){\tiny\color{magenta}1708${\text{-th}}$ frame}
            \end{overpic}
            \begin{overpic}[clip,width=0.1850\linewidth]{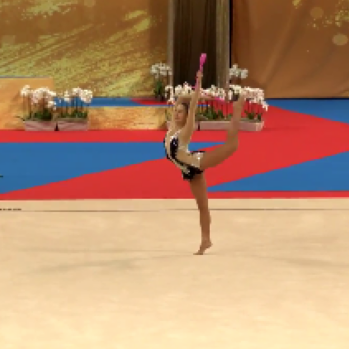}
                \put(3,5){\tiny\color{magenta}1714${\text{-th}}$ frame}
            \end{overpic}
            \begin{overpic}[clip,width=0.1850\linewidth]{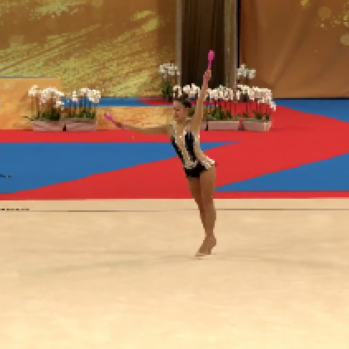}
                \put(3,5){\tiny\color{magenta}1720${\text{-th}}$ frame}
            \end{overpic}
            \\
            \multicolumn{2}{c}{\scriptsize(a) Sample \#056 from the RG (Club) dataset}
        \end{tabular}
    \end{minipage}
    \begin{minipage}{0.49\linewidth}
        \begin{tabular}{m{0.03\linewidth}<{\centering}m{0.97\linewidth}<{\centering}}
            \multicolumn{2}{c}{
                \begin{overpic}[clip,width=0.98\linewidth]{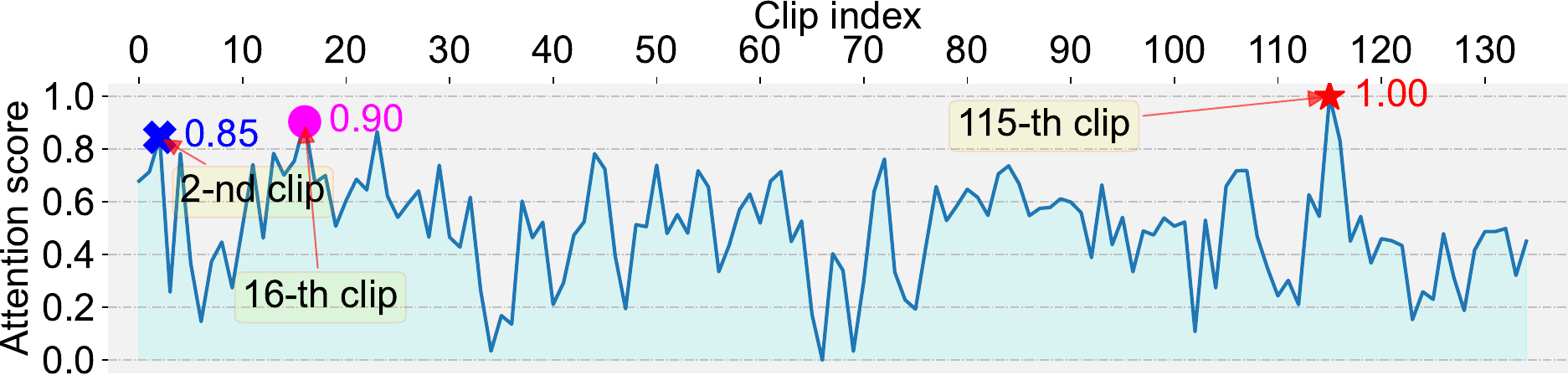}
                \end{overpic}
            }
            \\
            \rotatebox{90}{\scriptsize \color{magenta} 16${\text{-th}}$ clip}
             &
            \begin{overpic}[clip,width=0.1850\linewidth]{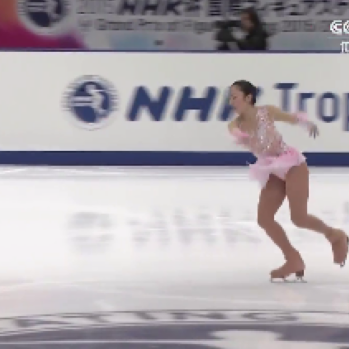}
                \put(3,5){\tiny\color{magenta}0511${\text{-st}}$ frame}
            \end{overpic}
            \begin{overpic}[clip,width=0.1850\linewidth]{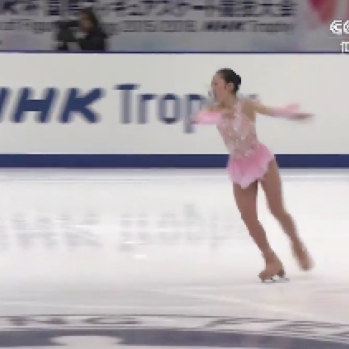}
                \put(3,5){\tiny\color{magenta}0516${\text{-th}}$ frame}
            \end{overpic}
            \begin{overpic}[clip,width=0.1850\linewidth]{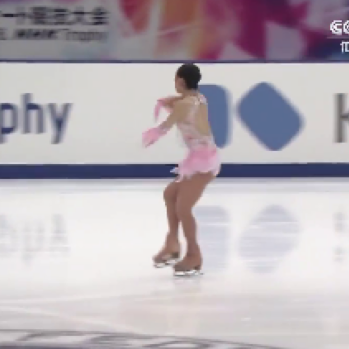}
                \put(3,5){\tiny\color{magenta}0524${\text{-th}}$ frame}
            \end{overpic}
            \begin{overpic}[clip,width=0.1850\linewidth]{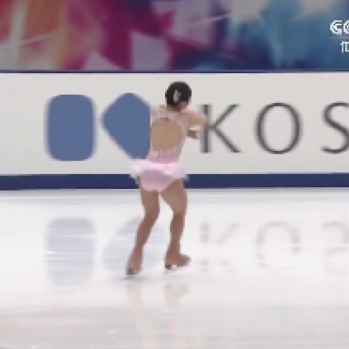}
                \put(3,5){\tiny\color{magenta}0530${\text{-th}}$ frame}
            \end{overpic}
            \begin{overpic}[clip,width=0.1850\linewidth]{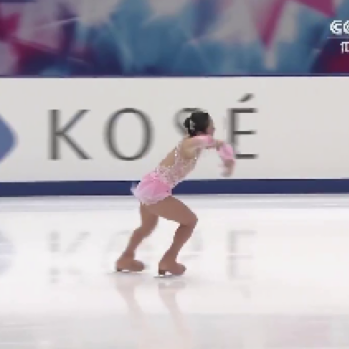}
                \put(3,5){\tiny\color{magenta}0536${\text{-th}}$ frame}
            \end{overpic}
            \\
            \rotatebox{90}{\scriptsize \color{red} 115${\text{-th}}$ clip}
             &
            \begin{overpic}[clip,width=0.1850\linewidth]{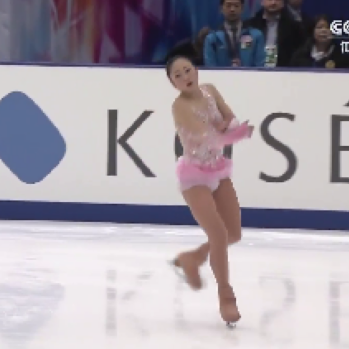}
                \put(3,5){\tiny\color{red}3680${\text{-th}}$ frame}
            \end{overpic}
            \begin{overpic}[clip,width=0.1850\linewidth]{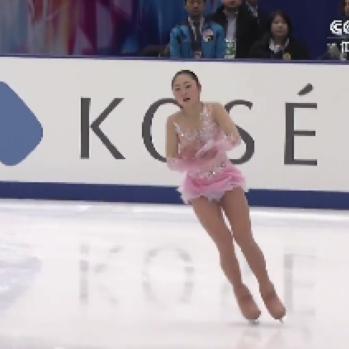}
                \put(3,5){\tiny\color{red}3686${\text{-th}}$ frame}
            \end{overpic}
            \begin{overpic}[clip,width=0.1850\linewidth]{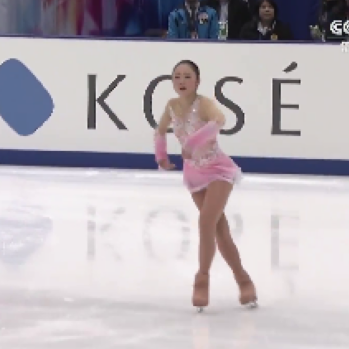}
                \put(3,5){\tiny\color{red}3691${\text{-st}}$ frame}
            \end{overpic}
            \begin{overpic}[clip,width=0.1850\linewidth]{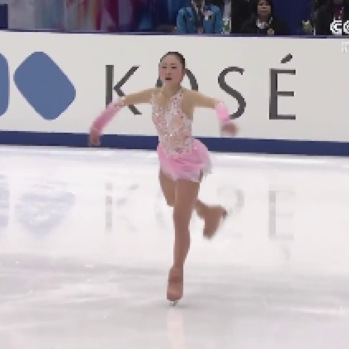}
                \put(3,5){\tiny\color{red}4698${\text{-th}}$ frame}
            \end{overpic}
            \begin{overpic}[clip,width=0.1850\linewidth]{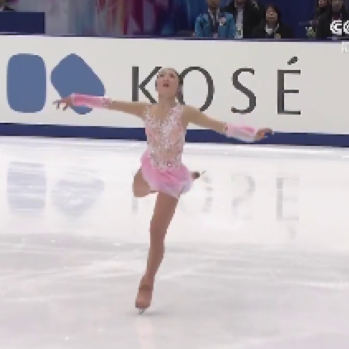}
                \put(3,5){\tiny\color{red}4704${\text{-th}}$ frame}
            \end{overpic}
            \\
            \rotatebox{90}{\scriptsize \color{blue} 116${\text{-th}}$ clip}
             &
            \begin{overpic}[clip,width=0.1850\linewidth]{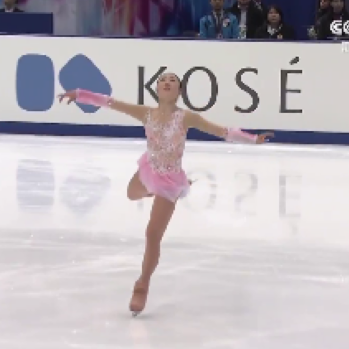}
                \put(3,5){\tiny\color{blue}4711${\text{-st}}$ frame}
            \end{overpic}
            \begin{overpic}[clip,width=0.1850\linewidth]{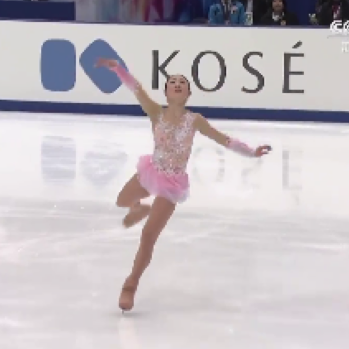}
                \put(3,5){\tiny\color{blue}4718${\text{-th}}$ frame}
            \end{overpic}
            \begin{overpic}[clip,width=0.1850\linewidth]{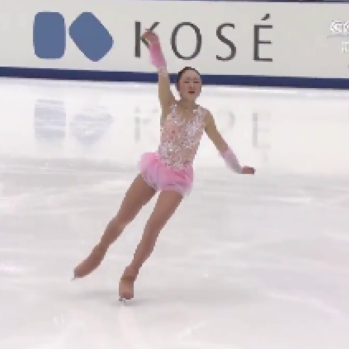}
                \put(3,5){\tiny\color{blue}4724${\text{-th}}$ frame}
            \end{overpic}
            \begin{overpic}[clip,width=0.1850\linewidth]{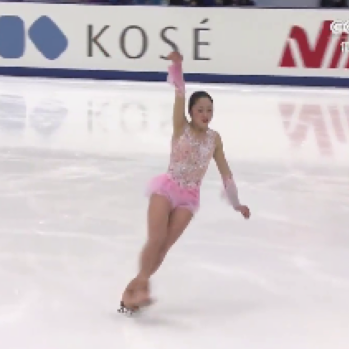}
                \put(3,5){\tiny\color{blue}4730${\text{-th}}$ frame}
            \end{overpic}
            \begin{overpic}[clip,width=0.1850\linewidth]{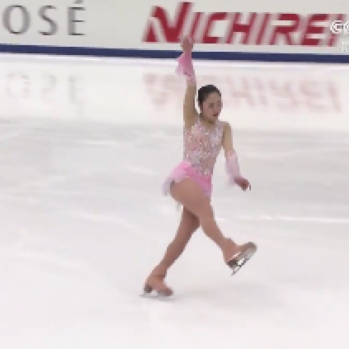}
                \put(3,5){\tiny\color{blue}4736${\text{-th}}$ frame}
            \end{overpic}
            \\
            \multicolumn{2}{c}{\scriptsize(b) Sample \#056 from the Fis-V (PCS) dataset}
        \end{tabular}
    \end{minipage}
    \caption{Visualization of attention weights and highlighted clips for samples on (a) the RG (Club) dataset and (b) the Fis-V (PCS) dataset.}
    \label{fig:att}
    {
        \phantomsubcaption\label{fig:att-a}
        \phantomsubcaption\label{fig:att-b}
    }
\end{figure*}

\begin{figure*}
    \centering
    \includegraphics[width=\linewidth]{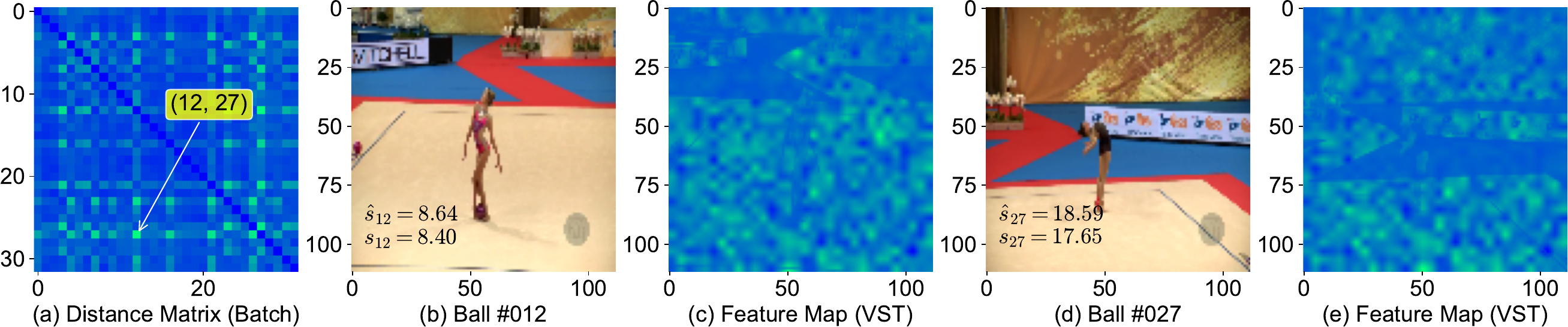}
    \caption{Visualization of the distance matrix on the Ball (RG) dataset.}
    {
        \phantomsubcaption\label{fig:distance-a}
        \phantomsubcaption\label{fig:distance-b}
        \phantomsubcaption\label{fig:distance-c}
        \phantomsubcaption\label{fig:distance-d}
        \phantomsubcaption\label{fig:distance-e}
    }
    \label{fig:distance}
\end{figure*}

\paragraph{Visualization of Attention Weights}
To gain deeper insights into the attention mechanism within the TETE module, we visualize the attention weights and the highlighted clips for two representative samples from the RG and Fis-V datasets.
\linelabel{r1:C1.6}\cref{fig:att-a,fig:att-b} illustrate the attention weight distributions for each clip in the Club (RG) and PCS (Fis-V) models, respectively.
The first row in each subfigure presents the normalized attention weights assigned to all action clips, revealing the varying levels of importance attributed to different segments of the action.
The following rows highlight the top three clips that received the highest attention scores, which are crucial for the model's decision-making.
In \cref{fig:att-a}, the model predicts a score of 14.30, while the ground-truth score is 14.52, resulting in a minimal error of 0.22. Similarly, in \cref{fig:att-b}, the model predicts a score of 27.78, compared to the ground-truth score of 26.68, yielding an error of 1.10. For example, in \cref{fig:att-b}, the three most attended clips are 39, 52, and 53, corresponding to key subactions where the player executes precise hand-leg coordination.
These visualizations provide valuable insights into how the AQA model prioritizes different temporal segments within an action sequence. By identifying the most critical moments, we enhance our understanding of the model’s interpretability and its evaluation process for long-term action assessment.

\paragraph{Visualization of the Distance Matrix}
In \cref{fig:distance}, the heatmap of the distance matrix is depicted. The highest value is located at $(12, 27)$, which signifies the greatest distance between the 12-th and 27-th actions. To further assess the efficacy, the two actions and their respective heatmaps are visualized. These heatmaps in \cref{fig:distance-c,fig:distance-d} provide insights into the domain shift issue. The minimal difference between the predicted and ground-truth scores indicates the effectiveness of our method. For instance, with a ground-truth score of $s_{12}=8.40$ and the predicted score is $\hat{s}_{12} = 8.64$. Additionally, observing the score distance between the two videos reveals a wide range, consistent with their feature distance, demonstrating the effectiveness of PHI.

\section{Conclusions and Discussions} \label{sec_conclusion}
In this work, we identify and analyze task-level and feature-level domain shifts in long-term AQA and propose PHI as a hierarchical adaptation framework to address feature-level discrepancies. Rather than a mere extension of existing methods, PHI introduces a novel integration of shallow-to-deep adaptation and coarse-to-fine alignment strategies to enhance AQA performance.
The shallow-to-deep adaptation strategy, enabled by GMF, effectively reduces domain gaps while maintaining computational efficiency. Simultaneously, the coarse-to-fine alignment mechanism, facilitated by LCR, refines coarse features extracted from pre-trained models, aligning them with fine-grained representations crucial for AQA. Experimental results on three representative long-term AQA datasets demonstrate the significant effectiveness of PHI, underscoring the importance of mitigating feature-level discrepancies in improving AQA performance.
Notably, compared to the task-level adaptation method CoFInAl, PHI exhibits superior performance in mitigating domain shifts and enhancing AQA accuracy, emphasizing the critical role of feature-level alignment in long-term AQA tasks.
\linelabel{r1:C2.1-d}Furthermore, the hierarchical adaptation framework of PHI is highly generalizable beyond AQA, with potential applications in rehabilitation analysis, sports motion scoring, and movement disorder diagnosis. The principles of shallow-to-deep adaptation and coarse-to-fine alignment also extend to domains such as multi-modal alignment. By addressing domain shifts through hierarchical feature refinement, PHI provides a theoretically grounded and versatile framework, paving the way for future research across multiple fields.

Despite its strong performance, PHI has several limitations, each suggesting directions for future work.
First, the auto-regressive nature of GMF introduces cumulative errors, which may progressively degrade prediction accuracy over multiple steps. Future research will focus on advanced optimization strategies and novel regularization techniques to mitigate these challenges.
\linelabel{r1:C2.3}Second, while our clip alignment method effectively minimizes discrepancies between actions, it may lead to information loss in cases of significant temporal variation. Future work will explore adaptive alignment strategies to enhance temporal correlation capture while maintaining computational efficiency.
\linelabel{r1:C3.6-b}Third, although PHI and CoFInAl address domain shifts from complementary perspectives, feature-level and task-level alignment, respectively, their integration into a unified framework remains an open challenge. Future research will investigate synergistic strategies to combine these approaches for a more comprehensive domain adaptation solution.
\linelabel{r1:C2.5-b}Additionally, PHI is currently designed for unimodal inputs, limiting its applicability in multi-modal scenarios. Extending the framework to support multi-modal integration will be a key avenue for future work, broadening its utility across diverse applications.
\linelabel{r1:C2.6}Finally, PHI is tailored for mitigating the domain shift issue in long-term AQA tasks, which may not be directly applicable to short-term AQA scenarios. Future research will explore adaptations to extend its applicability to short-term AQA, further increasing its versatility.
Addressing these limitations will enhance the applicability of PHI, advancing the field of AQA and its related domains.

\bibliographystyle{ieeetr}
\bibliography{refs}
\vspace{-3em}
\begin{IEEEbiography}[{\includegraphics[width=1in,height=1.1in,clip,keepaspectratio]{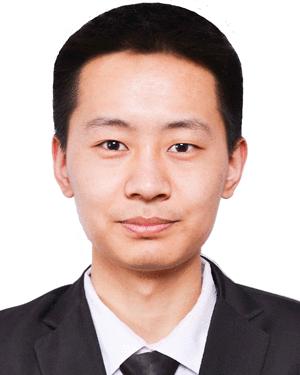}}]{Kanglei Zhou}
    is a Ph.D. candidate in the School of Computer Science and Engineering at Beihang University, specializing in action quality assessment and augmented reality. From February to August 2024, he was a visiting student in the Department of Computer Science at Durham University. He received his Bachelor's degree in the College of Computer and Information Engineering from Henan Normal University in 2020.
\end{IEEEbiography}
\vspace{-3em}
\begin{IEEEbiography}[{\includegraphics[width=1in,height=1.1in,clip,keepaspectratio]{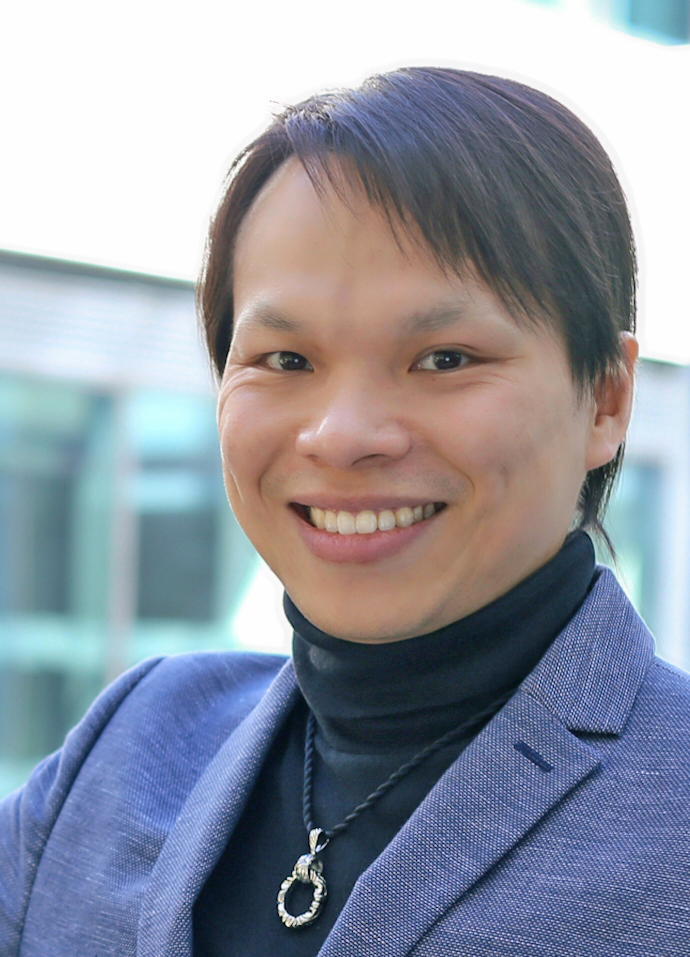}}]{Hubert P. H. Shum}
    (Senior Member, IEEE) is a Professor of Visual Computing and the Director of Research of the Department of Computer Science at Durham University, specialising in modelling spatio-temporal information with responsible AI. He is also a Co-Founder and the Co-Director of Durham University Space Research Centre. Before this, he was an Associate Professor/Senior Lecturer at Northumbria University and a Postdoctoral Researcher at RIKEN Japan. He received his PhD degree from the University of Edinburgh. He chaired conferences such as Pacific Graphics, BMVC and SCA, and has authored over 180 research publications.
\end{IEEEbiography}
\vspace{-3em}
\begin{IEEEbiography}[{\includegraphics[width=1in,height=1.1in,clip,keepaspectratio]{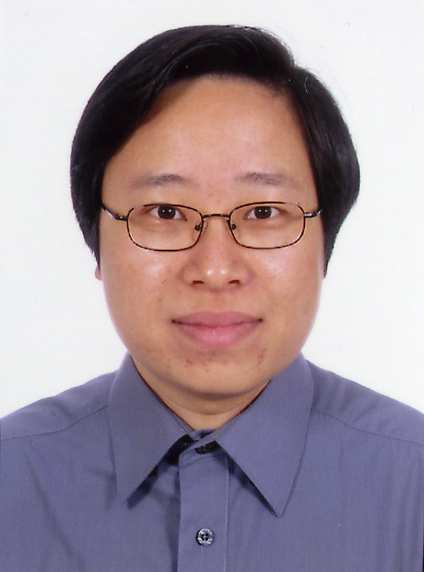}}]{Frederick W. B. Li} received a B.A. and an M.Phil. degree from Hong Kong Polytechnic University, and a Ph.D. degree from the City University of Hong Kong. He is currently an Associate Professor at Durham University, researching computer graphics, deep learning, collaborative virtual environments, and educational technologies. He is also an Associate Editor of Frontiers in Education and an Editorial Board Member of Virtual Reality \& Intelligent Hardware. He chaired conferences such as ISVC and ICWL.
\end{IEEEbiography}
\vspace{-3em}
\begin{IEEEbiography}[{\includegraphics[width=1in,height=1.1in,clip,keepaspectratio]{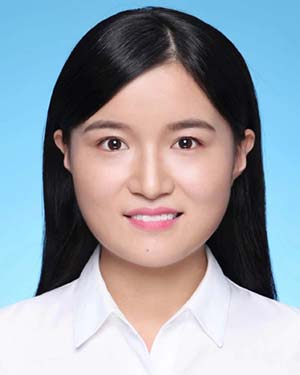}}]{Xingxing Zhang} received the BE and PhD degrees
    from the Institute of Information Science, Beijing
    Jiaotong University, in 2015 and 2020, respectively.
    She was also a visiting student with the Department
    of Computer Science, University of Rochester, from
    2018 to 2019. She was a postdoc with the Department
    of Computer Science and Technology, Tsinghua University, from 2020 to 2022. Her research interests include continual learning and zero/few-shot learning. She has received the excellent PhD thesis award from the Chinese Institute of Electronics, in 2020.
\end{IEEEbiography}
\vspace{-3em}
\begin{IEEEbiography}[{\includegraphics[width=1in,height=1.1in,clip,keepaspectratio]{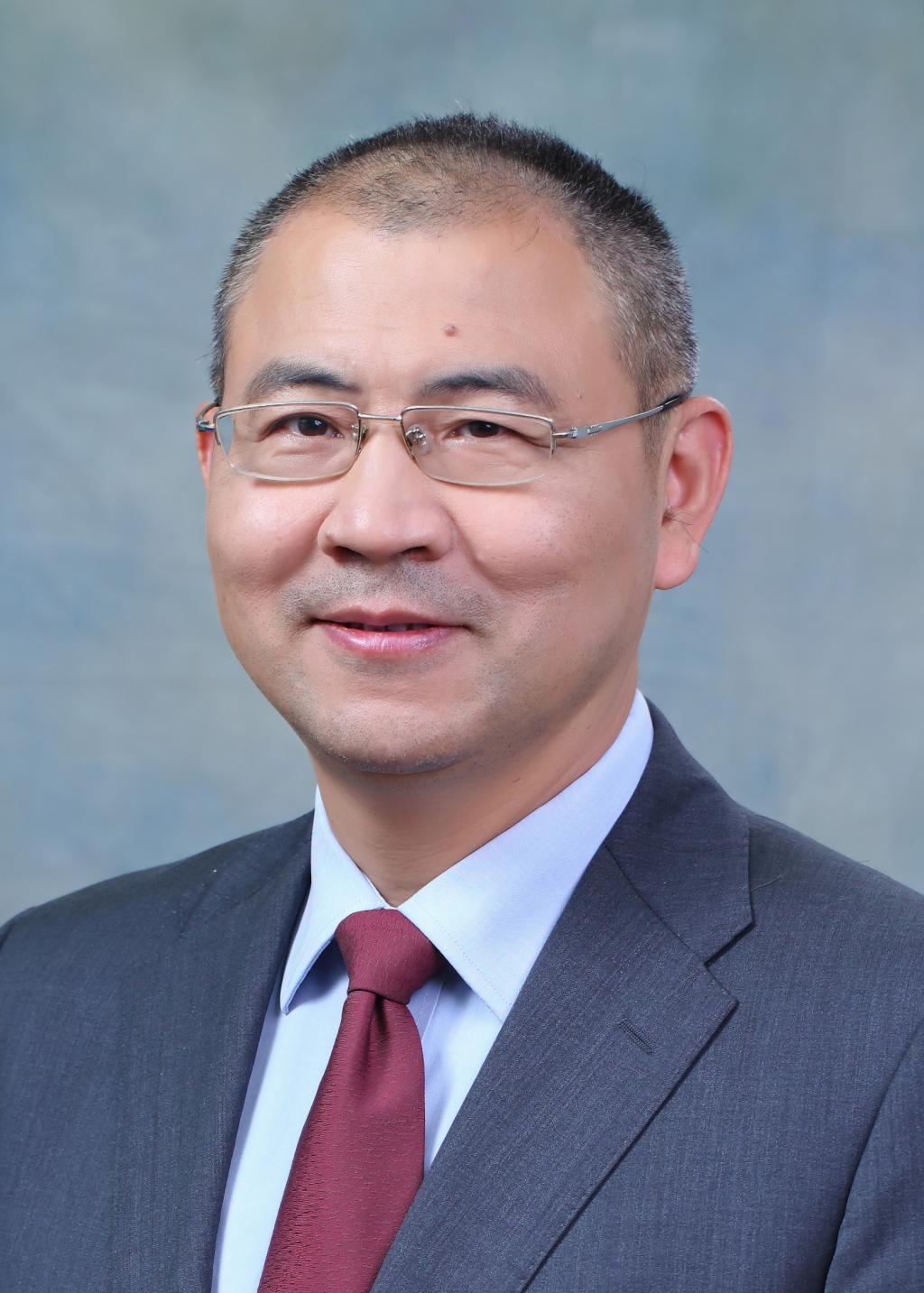}}]{Xiaohui Liang} received his Ph.D. degree in computer science and engineering from Beihang University, China. He is currently a Professor, working in the School of Computer Science and Engineering at Beihang University. His main research interests include computer graphics and animation, visualization, and virtual reality.
\end{IEEEbiography}
\vfill

\end{document}